\RequirePackage{fix-cm}
\documentclass[twocolumn]{svjour3}          
\usepackage[a4paper, margin=1in]{geometry}

\smartqed  
\usepackage{graphicx}
\usepackage{times}
\usepackage{epsfig}
\usepackage{amsmath}
\usepackage{amssymb}
\usepackage{subcaption}
\usepackage{bm}
\usepackage{booktabs}
\usepackage{multirow}
\usepackage{makecell}
\usepackage{array}
\usepackage{cite}
\usepackage{color}
\usepackage[english]{babel}
\usepackage[misc]{ifsym}
\usepackage{blindtext}
\usepackage{natbib}
\usepackage[ruled]{algorithm2e}
\usepackage{indentfirst}
\usepackage[switch]{lineno}
\usepackage{enumitem}
\definecolor{graycolor}{rgb}{0.95,0.95,0.95}
\usepackage{bbding}
\usepackage{pifont}
\usepackage{blindtext}
\usepackage{algorithmic}
\usepackage{listings}
\usepackage{colortbl}
\usepackage{varioref}
\usepackage[backref]{hyperref} 
\hypersetup{hidelinks}
\usepackage{microtype}

\def\ie{\emph{i.e.}} 

\def\eg{\emph{e.g.}}

\def\etc{\emph{etc.}}

\def\etal{\emph{et al. }}

\newcommand{\Tref}[1]{Table~\ref{#1}}

\newif\ifthickvlines
\newcolumntype{|}{!{
	\ifthickvlines
	\vrule width1.5\arrayrulewidth
	\else
	\vline
	\fi}}

\definecolor{mypink}{rgb}{.99,.91,.95}

\hypersetup{
    colorlinks=true,
    linkcolor=blue,
    filecolor=blue,      
    urlcolor=blue,
	citecolor=blue,
}

 \journalname{IJCV}
\begin{document}

\title{Unaligned RGB Guided Hyperspectral Image Super-Resolution with Spatial-Spectral Concordance
}


\author{Yingkai Zhang$^1$         \and
        Zeqiang Lai$^1$ 	\and
        	Tao Zhang$^2$ \and
		Ying Fu$^{1}$  \and 
		Chenghu Zhou$^{3,4}$
}



\institute{
	Yingkai Zhang\\
	\hspace*{3mm} E-mail: zhangyingkai@bit.edu.cn\\
$\textrm{\Letter}$ Ying Fu\\
	\hspace*{3mm} E-mail: fuying@bit.edu.cn\\ \\
	\\
	$1$ School of Computer Science and Technology, Beijing Institute of Technology, Beijing, 100081, China\\ \\
	$2$ School of Communication Engineering, Hangzhou Dianzi University, Hangzhou, 310018, China \\ \\
	$3$ State Key Laboratory of Resources and Environmental Information System, Institute of Geographic Sciences and Natural Resources Research, Chinese Academy of Sciences, Beijing 100101, China \\ \\
	$4$ College of Resources and Environment, University of Chinese Academy of Sciences, Beijing 100049, China \\ \\
 }

\date{Received: date / Accepted: date}

\maketitle
\setlength{\parindent}{1em}

\begin{abstract}

Hyperspectral images (HSIs) super-resolution (SR) aims to improve the spatial resolution, yet its performance is often limited at high-resolution ratios. 
The recent adoption of high-resolution reference images for super-resolution is driven by the poor spatial detail found in low-resolution HSIs, presenting it as a favorable method.
However, these approaches cannot effectively utilize information from the reference image, due to the inaccuracy of alignment and its inadequate interaction between alignment and fusion modules.
In this paper, we introduce a \textbf{S}patial-\textbf{S}pectral \textbf{C}oncordance \textbf{H}yperspectral \textbf{S}uper-\textbf{R}esolution (SSC-HSR) framework for unaligned reference RGB guided HSI SR to address the issues of inaccurate alignment and poor interactivity of the previous approaches.
Specifically, to ensure spatial concordance, \ie, align images more accurately across resolutions and refine textures, we construct a Two-Stage Image Alignment (TSIA) with a synthetic generation pipeline in the image alignment module, where the fine-tuned optical flow model can produce a more accurate optical flow in the first stage and warp model can refine damaged textures in the second stage.
To enhance the interaction between alignment and fusion modules and ensure spectral concordance during reconstruction, we propose a Feature Aggregation (FA) module and an Attention Fusion (AF) module. 
In the feature aggregation module, we introduce an Iterative Deformable Feature Aggregation (IDFA) block to achieve significant feature matching and texture aggregation with the fusion multi-scale results guidance, iteratively generating learnable offset. 
Besides, we introduce two basic spectral-wise attention blocks in the attention fusion module to model the inter-spectra interactions.
Extensive experiments on three natural or remote-sensing datasets show that our method outperforms state-of-the-art approaches on both quantitative and qualitative evaluations.
Our code will be publicly available to the community.
\end{abstract}

\begin{figure}
\begin{center}
\includegraphics[width=1\linewidth]{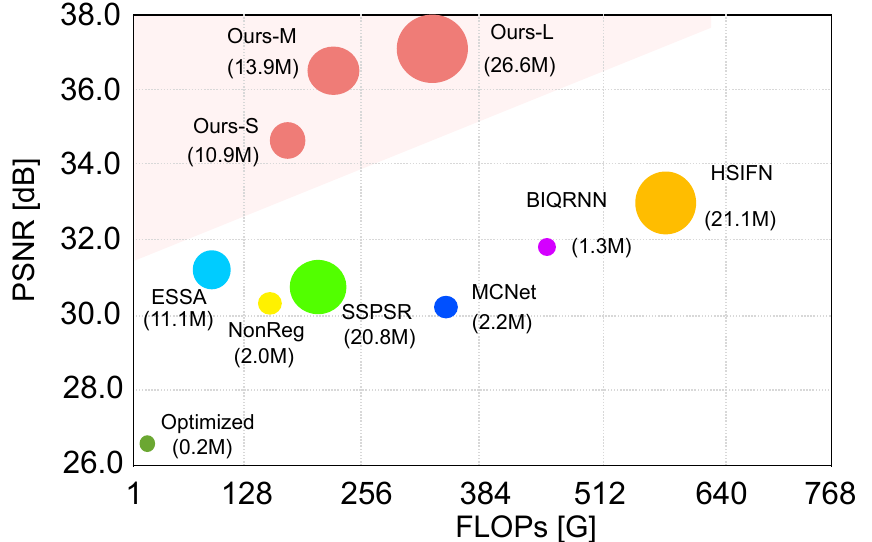}
\end{center}
   \caption{PSNR-Params-FLOPs comparisons with other methods. The vertical coordinate indicates PSNR (dB), the horizontal coordinate indicates FLOPs (computational cost), and the size of the bubbles indicates Params (memory cost)}
\label{fig:psnr_ssim_params_flops}
\end{figure}

\section{Introduction}

Hyperspectral image (HSI) contains abundant spectral information that enables a variety of applications \citep{azar2020hyperspectral, zhang2010object, pan2003face, kim20123d, zhang2022guided, su2023making, huynh2013shape, fu2023category, li2024supervise}, but hyperspectral cameras often sacrifice their spatial resolution for high spectral resolution \citep{funatomi2022eliminating}. Thus, a series of researches have been proposed in recent years to solve this problem. 
One line of research has focused on restoring high-resolution (HR) HSI from a single low-resolution (LR) HSI \citep{li2020mixed, fu2021bidirectional, zhang2023essaformer, Guo_2023_CVPR}. Despite certain effectiveness, these methods are still limited by their less pleasant restoration results from single LR input HSI, especially for large resolution gaps, which cannot provide more meaningful information. 

To address the limitation of the single HSI super-resolution (SR), an alternative line of research has emerged, referred to as fusion-based HSI SR \citep{fu2019hyperspectral, Guo_2023_CVPR, wu2023hsr, hu2022hyperspectral}, which employs additional more accessible HR RGB or multispectral image (MSI) as references. All of the above fusion-based methods are based on the strict assumption of precise alignment, which can greatly affect the performance of the fusion model. However, obtaining strictly aligned images can be challenging in the real world, which limits the practical use of these fusion-based methods. 

\begin{figure}
\centering
\includegraphics[width=1\linewidth]{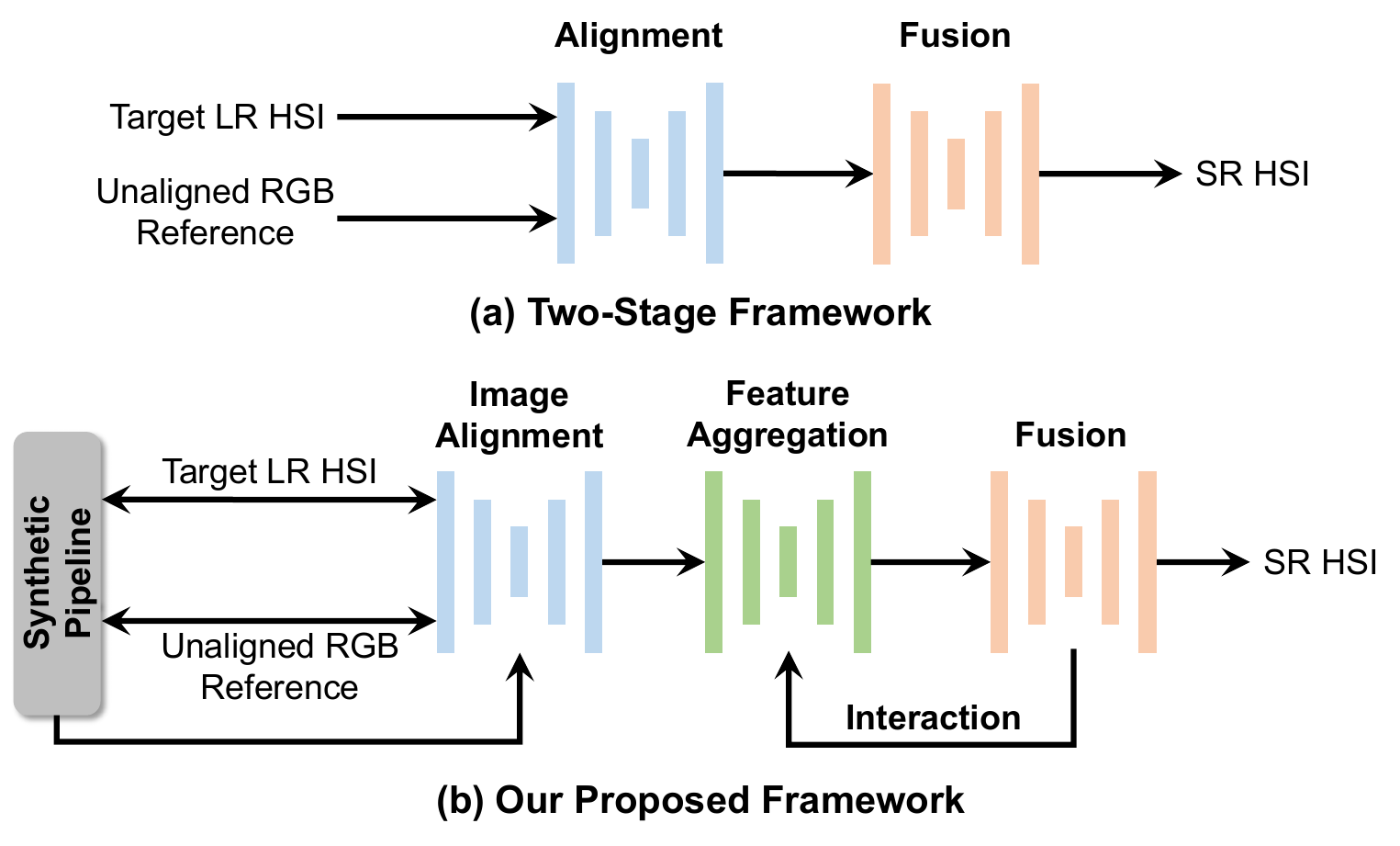}
   \caption{Illustration of two unaligned reference-based hyperspectral image super-resolution frameworks. (a) Two-stage framework. (b) The proposed framework of Ours}
\label{fig:different_framework}
\end{figure}

Recently, some work \citep{fu2020simultaneous, zheng2022nonregsrnet, lai2024hyperspectral, ying2022unaligned} use unaligned simulated or real data to improve practical applications and specifically alleviate the assumption of precise alignment, and further by designing specific method for HSI SR from different perspectives. 
Although these approaches have improved their practicality, they still have several shortcomings.
First, most previous works \citep{zhou2019integrated, ying2022unaligned} utilize a separate two-stage framework shown in Fig.~\ref{fig:different_framework}(a), which means that the fusion model relies heavily on the alignment module. Inaccurate alignment can accumulate errors, resulting in low-quality restored HSI. Besides, pairs of data exist in resolution gap and distribution gap, which seriously affects the effectiveness of matching between the data.
Second, although previous works \citep{zheng2022nonregsrnet, lai2024hyperspectral} have been conducted on the interaction between alignment and fusion through a single-step manner, they could not learn effective spatial transformations, and do not adequately perform spatial and spectral information aggregation, which also lead to ambiguous restored results.

In this paper, we mainly focus on LR HSI super-resolution guided by unaligned reference HR RGB image. To address the above two issues, we propose the spatial-spectral concordance hyperspectral super-resolution (SSC-HSR) network. As shown in Fig.~\ref{fig:different_framework}(b), our proposed framework mainly has three parts, image alignment with a synthetic pipeline, feature aggregation, and attention fusion. 
To improve the alignment accuracy of LR HSI and HR RGB across resolutions (\ie, resolution gaps) and distributions (\ie, distribution gaps), we first introduce an image alignment module based on the two-stage alignment model and a synthetic generation pipeline. Due to the lack of real cross-resolution data pairs with accurate optical flow between HSI and RGB, which denotes the transformational relationships between target and reference images, \ie, the pixel-wise correspondence. Thus, we adopt a synthetic pipeline to generate the data with accurate optical flow. 
Based on these data, we introduce a two-stage image alignment module to fine-tune the optical flow model in the first stage and train a WarpNet in the second stage for better refinement of textures.

To enhance the interaction of fine-grained feature aggregation and fusion modules, and further aggregate efficient global similar features, we present an iterative deformable feature aggregation (IDFA) block. It is based on the global attention matching (GAM) block, which aims to capture the global relevant position between features and generate their initialization weights prior to IDFA. We utilize this knowledge to initialize the offset and mask of the deformable convolution, the component of IDFA, and then iteratively combine it with the output of the fusion block to generate a learnable offset and mask. We can learn finer-grained spatial offsets and aggregation of the most similar features by combining the above knowledge.
In the attention fusion module, we utilize two basic spectral-wise attention blocks guided by extracted multi-scale features, which help to model the dependence of spectral-wise channels between features for guaranteeing the spectral concordance and circumvents the high spatial sparsity of HSIs.

We conduct our method SSC-HSR on three hyperspectral datasets, containing natural and remote-sensing scenes. Quantitative and visual results demonstrate that our method achieves state-of-the-art (SOTA) performance, especially at high resolutions where it still maintains superior performance.
The main contributions of our work are summarized as follows:

\begin{itemize}
    \item We propose a network, named SSC-HSR, for unaligned RGB guided hyperspectral super-resolution, which can well handle joint of alignment and super-resolution. Both quantitative and qualitative experiments on three hyperspectral datasets demonstrate the performance of our method.
    \item We introduce a two-stage image alignment process complemented by a synthetic generation pipeline. It first fine-tunes the optical flow model, ensuring the generation of precise optical flow. Subsequently, it employs a warp model aimed at refining the image texture details, taking into account the cross-resolution and distribution disparities of the data.
    \item We introduce an iterative deformable feature aggregation block in the feature aggregation module. This block is designed to learn finer-grained spatial offsets and relevant feature aggregation through progressively enhanced interaction with the fusion model. Additionally, the fusion model employs reference-guided attention to model spectral-wise dependencies, ensuring spectral concordance.
\end{itemize}

\section{Related Work}
\label{sec:related-work}

In this section, we review the most relevant studies on single HSI super-resolution, fusion-based HSI super-resolution, and unaligned reference-based HSI super-resolution.

\subsection{Single HSI Super-Resolution}

The realm of Single Image Super-Resolution (SISR), particularly within the hyperspectral imaging spectrum, presents a complex and multifaceted challenge that has garnered substantial interest in recent years. The fundamental complexity arises from the inherent ambiguity in the process: a single low-resolution (LR) HSI could originate from a multitude of high-resolution (HR) HSIs. 
Thanks to advancements in deep learning for low-level image processing \citep{mei2017hyperspectral, 2024deep, 2022gan, 2024freqfusion, 2023instance, sun2020hyperspectral}, this issue has spurred a surge of research utilizing deep convolutional neural network (CNN) methodologies for HSI reconstruction.
This issue has prompted a surge in research employing deep convolutional neural network (CNN) methodologies for HSI reconstruction. 
A plethora of studies \citep{hu2020hyperspectral, fu2022joint, fu2022coded, jiang2020learning, li2020mixed, fu2021bidirectional, zhang2023essaformer, fu2017adaptive} have made significant strides in this direction.

A notable contribution by Mei \etal \citep{mei2017hyperspectral} introduces an innovative three-dimensional full CNN, adept at harnessing both spatial and spectral data. Sun \etal \citep{sun2020hyperspectral} adopt a feature pyramid for multi-scale feature extraction of hyperspectral images. Further, Jiang \etal \citep{jiang2020learning} propose a novel method for learning spatial-spectral priors using group convolution. Mixed 2D and 3D convolutions are explored by Li \etal \citep{li2020mixed}, while Fu \etal \citep{fu2021bidirectional} introduces a bidirectional 3D quasi-recurrent neural network (BiQRNN) to delve into spectral correlations.

With the advent of transformer in image processing \citep{dosovitskiy2020image, liu2021swin, li2024latent, tian2023transformer}, a new frontier has opened in HSI SISR, as explored by Hu \etal \citep{hu2021hyperspectral} and Liu \etal \citep{liu2022interactformer}. ESSAformer \citep{zhang2023essaformer} introduces the spectral correlation coefficient of the spectrum and a specific self-attention mechanism to enlarge the receptive field without much more computation. Nonetheless, the scarcity of HSI training samples, coupled with the absence of spatial high-frequency information in HSI, poses significant challenges in achieving superior super-resolution for LR HSI without auxiliary data.

\begin{figure*}
\centering
\includegraphics[width=1\linewidth]{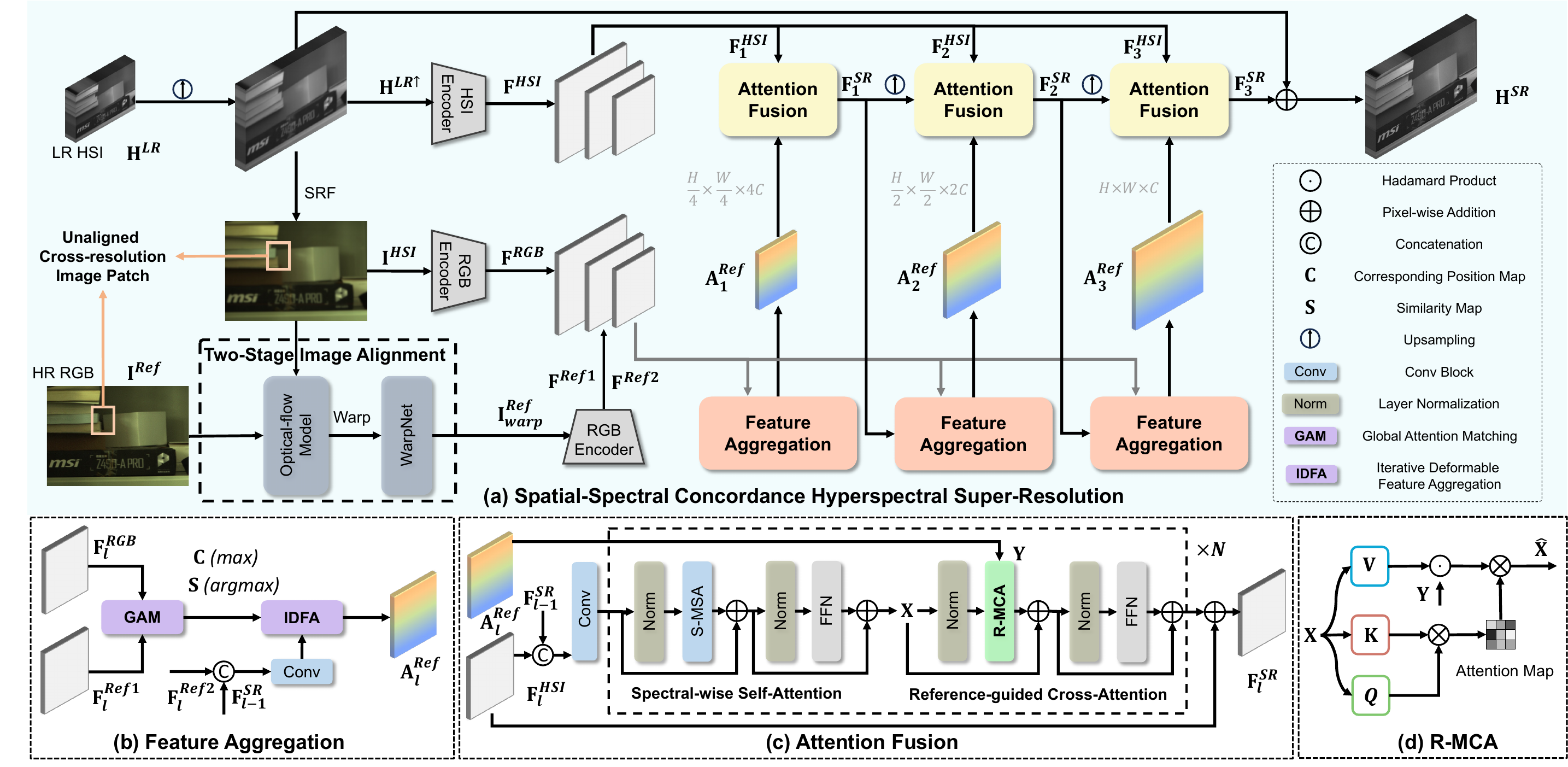}
   \caption{Overall framework (a) of SSC-HSR. Our method is mainly composed of three components, \ie, 
   Two-Stage Image Alignment module includes an optical flow model to predict optical flow for RGB reference warping, a based on optical flow to warp RGB reference, and a warp model (WarpNet) to refine the image texture. 
   Feature Aggregation module includes Global Attention Matching (GAM) and Iterative Deformable Feature Aggregation (IDFA) to achieve feature matching and texture aggregation iteratively.
   Attention Fusion module based on spectral-wise self-attention (SSA) block and reference-guided cross-attention (RCA) block to explore spectral correspondence and finally fuse the features of HSI and RGB reference. 
   The illustrations of feature aggregation and attention fusion are shown as (b) and (c) respectively. (d) is the brief details of reference-guided multi-head cross-attention (R-MCA). The architecture of spectral-wise multi-head self-attention (S-MSA) likes to R-MCA but without input $\mathbf{Y}$}
\label{fig:framework}
\end{figure*}

\subsection{Fusion-based HSI Super-Resolution}

In order to address the inherent problem of lack of high-frequency detail information in a single image in the single HSI super-resolution methods, fusion-based approaches use an auxiliary HR RGB/multispectral image (MSI) to provide high-frequency texture details. 
Most existing fusion-based works consider using a strictly aligned reference image as guidance for HSI SR. The aligned input data, LR HSI and HR RGB/MSI, are always simulated from a single HR HSI, \ie, spatial downsampling and spectral downsampling, respectively.

Some works \citep{akhtar2014sparse, akhtar2015bayesian, akhtar2016hierarchical, dong2016hyperspectral, kawakami2011high, li2018fusing, zhang2018exploiting} apply different optimization priors, while others \citep{fu2019hyperspectral, qu2018unsupervised, dian2018deep, hu2022hyperspectral, Guo_2023_CVPR, wu2023hsr, zhou2023memory, wang2023general, zhang2024rgb} are based on deep CNN or Transformer. Akhtar \etal use sparse representation to exploit signal sparsity and non-negativity in \citep{akhtar2014sparse}, and non-parametric Bayesian sparse representation to learn Bayesian dictionaries in \citep{akhtar2015bayesian}. Kawakami \etal \citep{kawakami2011high} search a matrix factorization for the unmixing algorithm based on the spatial sparsity of HSI. Li \etal \citep{li2018fusing} fuse input images based on coupled sparse tensor factorization. Dian \etal \citep{dian2018deep} utilize a prior learned by deep CNN to incorporate into an optimized fusion framework. Fu \etal \citep{fu2019hyperspectral} present a simple and efficient CNN instead of using hand-crafted priors in an unsupervised way and set up a real hybrid camera system with a beam splitter for two optical paths to obtain image pairs. Yet, the intensity of each optical path will be reduced and the result is very dependent on the design of the hardware, otherwise, the wavefront information (\ie, the phase and shape of the light waves) can be corrupted due to imperfect design.
DAEM \citep{Guo_2023_CVPR} develop a coordination optimization HSI super-resolution framework, which can establish a cycle between the fusion model and explicit degradation estimation with predicted different blur kernels. HSR-Diff \citep{wu2023hsr} introduces an approach based on conditional diffusion models to generate an HR-HSI via refinement iteratively.
However, these works are based on well-alignment, which is highly dependent on complex imaging systems and precise calibration processes. As a result, these seriously undermine the usefulness of such approaches.

\subsection{Unaligned Reference-based HSI Super-Resolution}

Previous fusion-based methods have been challenging to employ in real-world scenarios due to their reliance on strict alignment priors. To mitigate this limitation and enhance the practicality of the designed approach, some work \citep{fu2020simultaneous, nie2020unsupervised, qu2022unsupervised, zheng2022nonregsrnet, zhou2019integrated, ying2022unaligned} generate unaligned simulation data through rigid or non-rigid transformation. Fu \etal \citep{fu2020simultaneous} presents an approach to align images by learning geometric parameters through minimal optimization. Nie \etal \citep{nie2020unsupervised} also learn the parameters of the affine transformation via a spatial transformer network in an unsupervised way. Qu \etal \citep{qu2022unsupervised} propose an unsupervised network to learn spectral and spatial information by maximizing the mutual information from the two modalities. It is a single-stage, \ie end-to-end model to avoid explicit alignment. Zhou \etal \citep{zhou2019integrated} present an integrated registration and fusion approach, which has shown superiority in several datasets. NonRegSRNet \citep{zheng2022nonregsrnet} uses a spatial transformer network for alignment. HSIFN \citep{lai2024hyperspectral} collects a real dataset with more complex transformations and presents a method to obtain HR HSI using optical flow based on the pre-trained model for alignment.
However, most of them are limited to precise spatial alignment while overlooking the issue of spectral misalignment. They are significant in modeling the spatial-spectral correspondence of HSI and RGB reference.
In this work, we propose SSC-HSR for unaligned RGB guided HSI super-resolution, spatially global matching and aggregating related features, and exploring the spectral correspondence between HSI and RGB reference.

\section{Method}
\label{sec:method}

In this section, we first formulate the problem for unaligned RGB guided HSI super-resolution and describe the motivation for our proposed method in Section~\ref{sec: formulation and motivation}. Then, we introduce our two-stage image alignment in Section~\ref{sec: two-stage image alignment}. Next, we mainly show the details about our method of feature extraction, feature aggregation, and attention fusion in Sections~\ref{sec: feature extraction},~\ref{sec: feature aggregation}, and~\ref{sec: attention fusion}, respectively.
The overview of the proposed framework is shown in Fig.~\ref{fig:framework}.

\begin{figure}
\begin{center}
\setlength{\tabcolsep}{0.01cm}
    \begin{subfigure}[h]{0.15\textwidth}
          \centering
          \small
          \includegraphics[width=1\linewidth]{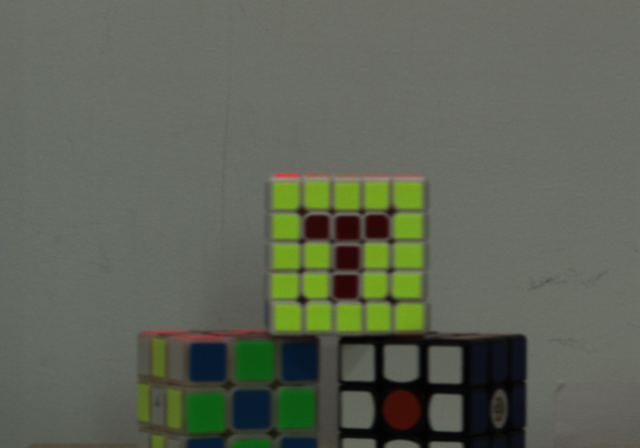}
          \caption{Target}
      \end{subfigure}
    \hfill
    \begin{subfigure}[h]{0.15\textwidth}
          \centering
          \small
          \includegraphics[width=1\linewidth]{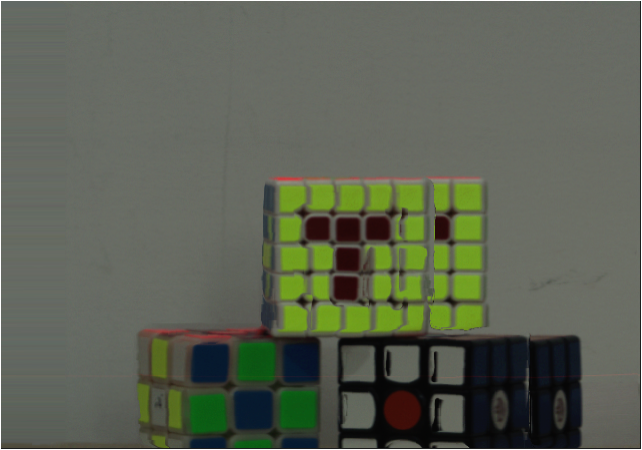}
          \caption{RAFT w/o fine-tuned}
      \end{subfigure}
    \hfill
    \begin{subfigure}[h]{0.15\textwidth}
          \centering
          \small
          \includegraphics[width=1\linewidth]{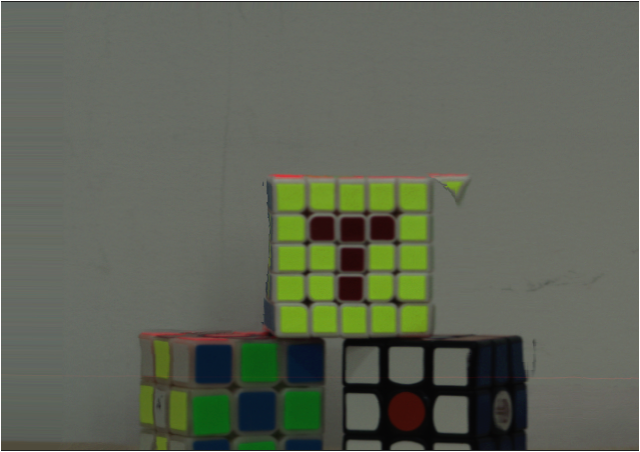}
          \caption{RAFT w/ fine-tuned}
      \end{subfigure}
    \vfill
    \begin{subfigure}[h]{0.15\textwidth}
          \centering
          \small
          \includegraphics[width=1\linewidth]{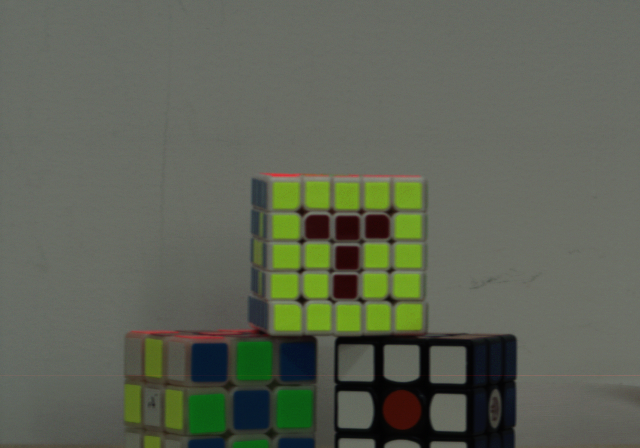}
          \caption{Reference}
      \end{subfigure}
    \hfill
    \begin{subfigure}[h]{0.15\textwidth}
          \centering
          \small
          \includegraphics[width=1\linewidth]{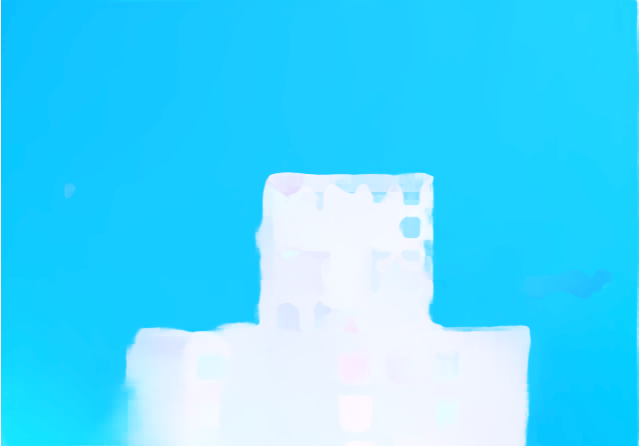}
          \caption{Optical Flow}
      \end{subfigure}
    \hfill
    \begin{subfigure}[h]{0.15\textwidth}
          \centering
          \small
          \includegraphics[width=1\linewidth]{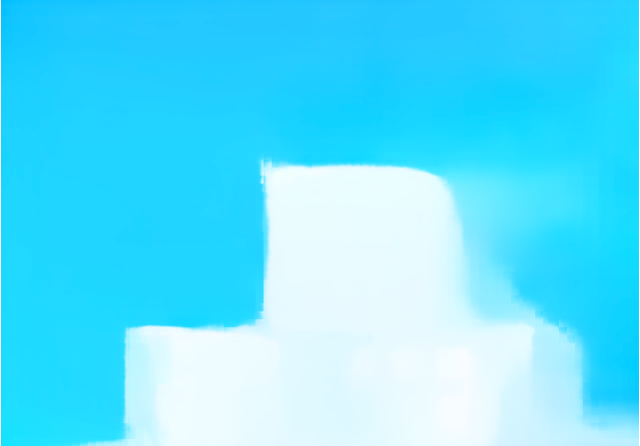}
          \caption{Optical flow}
      \end{subfigure}
\end{center}
   \caption{Optical flow prediction results at $\times8$ cross-resolution of LR HSI and HR RGB. (a) The target HSI. (b) The warped image by RAFT pre-trained on the FlyingChairs and FlyingThings datasets. (c) The warped image by RAFT fine-tuned with synthesized data from our generation pipeline. (d) The reference image. (e) The optical flow predicted by RAFT without fine-tuning. (f) The optical flow predicted by RAFT with fine-tuned}
    \label{fig:optical-flow}
\end{figure}

\begin{figure*}
\begin{center}
 \includegraphics[width=1\linewidth]{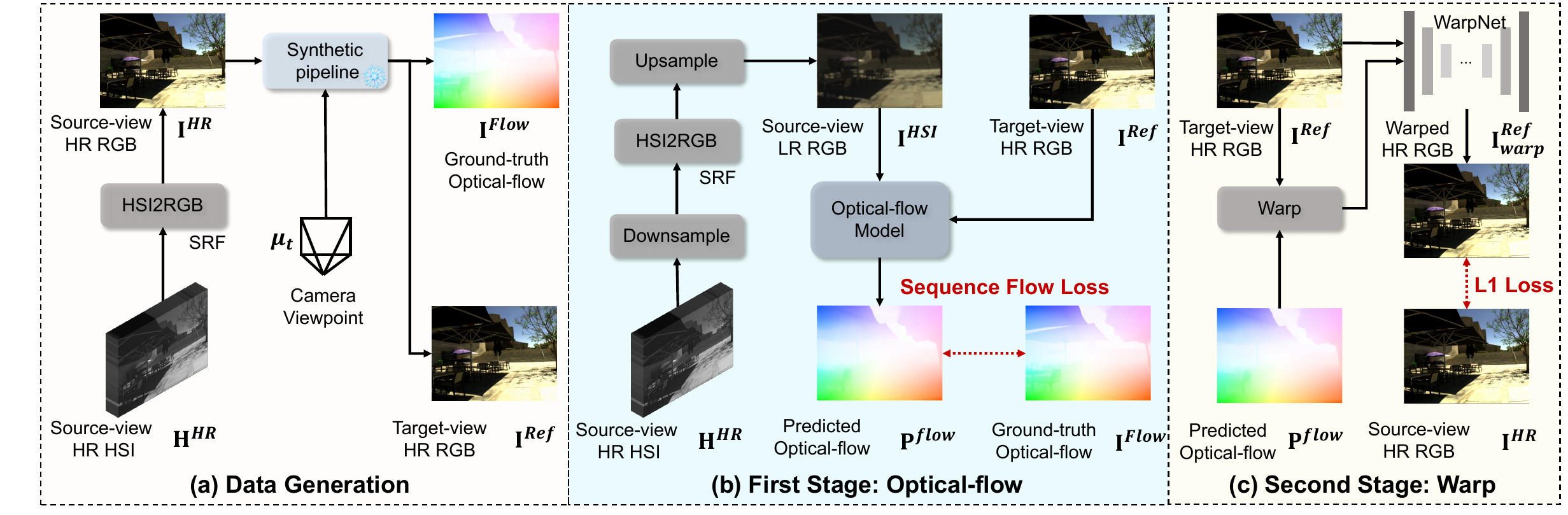}
\end{center}
   \caption{The illustration of the \textbf{two-stage image alignment} with the synthetic generation pipeline. The snowflake icon on the synthetic pipeline indicates that the models used in this part are frozen and not involved in training. Image alignment modules mainly contain an optical flow model and a warp model (\ie, WarpNet), which are trained in the first stage and second stage, respectively}
\label{fig:modules}
\end{figure*}

\subsection{Formulation and Motivation.}
\label{sec: formulation and motivation}
Unaligned RGB guided HSI SR aims to produce an HR HSI from LR HSI and unaligned HR RGB reference.
Suppose that given a LR HSI $\mathbf{H}^{LR}\in\mathbb{R}^{h\times w\times B}$ and an unaligned HR RGB reference image $\mathbf{I}^{Ref}\in\mathbb{R}^{H\times W\times b}$.  $H$, $W$, and $B$ are the number of rows, columns, and bands of spectrum for the high-resolution. Correspondingly, $h$, $w$, and $b$ denote the low-resolution representations. We aim to produce a high-quality HR HSI, which can be expressed as
\begin{equation}
	\mathbf{H}^{SR} = f(\mathbf{H}^{LR}, \mathbf{I}^{Ref}\vert\theta),
\end{equation}
where $\mathbf{H}^{SR}\in\mathbb{R}^{H\times W\times B}$ denotes the HR HSI, $f(\cdot\vert\theta)$ represents our network, and $\theta$ is the parameters of the network.
Some previous works have sought to alleviate the assumption of precise alignment, but they still face several shortcomings limiting their real application. 

On the one hand, as shown in Fig.~\ref{fig:different_framework}, most previous works like Integrated \citep{zhou2019integrated}, NED \citep{ying2022unaligned} utilize a two separate stage framework, containing alignment model and fusion model in Fig.~\ref{fig:different_framework}(a). The super-resolution results of the fusion stage are based on the aligned output of the alignment stage. This requires that the alignment must have a high degree of precision, otherwise, it can accumulate errors leading to low-quality results. HSIFN \citep{lai2024hyperspectral} is based on the optical flow model to warp multi-scale features. 
Nevertheless, the generic pre-trained model falls short of effectively addressing the issue of image alignment when faced with disparities in resolution and distribution.
As shown in Fig.~\ref{fig:optical-flow}, the warped image with the original pre-trained model on the FlyingChairs \citep{dosovitskiy2015flownet} and FlyingThings \citep{mayer2016large} datasets is obviously more redundancy or destroyed texture compared with the warped image generated by the fine-tuned model in our pipeline.
 
On the other hand, some works adopt the two-stage framework shown in Fig.~\ref{fig:different_framework}(a) in a fusion manner, \ie, the results of the alignment model and fusion model will affect each other to assist alignment and super-resolution tasks. However, previous works suffer from insufficient interaction and failure to capture clean features. NonReg \citep{zheng2022nonregsrnet} assemble the registration module and fusion module into a unified network, which can leverage mutual reinforcement across multitasking. Yet, it can be regarded as a single-step fusion manner whose insufficient interaction makes it difficult to effectively utilize two different image information and often produces low-quality results. HSIFN \citep{lai2024hyperspectral} proposes a multi-stage pixel-wise alignment module and attention module to address the misalignment module. 
Due to inaccurately learned spatial offsets and limited interaction between the alignment and fusion modules, the presence of multi-scale processing may still result in blurred textures, among other issues, adversely affecting the quality of reconstruction results, which is proven in Section~\ref{sec: ablation}.

Thus, to generate a more accurate optical flow based on the optical flow model and refine the texture from damaged regions, we propose a two-stage image alignment module and a synthetic generation pipeline. To more robustly aggregate features and enhance the interaction of alignment and fusion modules, we propose an iterative deformable feature aggregation block. To model the correspondence of featured spectral channels, we introduce an attention mechanism combined with spectral-wise self-attention and reference-guided cross-attention blocks. 

\subsection{Two-Stage Image Alignment}
\label{sec: two-stage image alignment}

From Fig.~\ref{fig:optical-flow}, we observe that an inaccurate image alignment module leads to broken textures, which can produce even worse results than directly using unaligned images as input to an end-to-end network, especially the data pairs with resolution and distribution gaps. Thus, we propose a more robust Two-Stage Image Alignment (TSIA) in this section, which mainly has two components, a synthetic pipeline for data generation and image alignment models.

Optical flow models have an important role in several RGB domain tasks, low-light enhancement \citep{zhang_2021_CVPR}, super-resolution \citep{tan2020crossnet++}, \etc However, these works mainly utilize a pre-trained optical flow model to predict pixel-wise offsets, which is not suitable for image pairs with resolution and distribution gaps between LR HSI and HR RGB in our task. Even though multi-scale is used to eliminate the effects of the gaps, it still affects the overall model performance. Thus, we introduce a synthetic generation pipeline for further training the optical flow model and WarpNet. 

Based on the RGB prior knowledge, we first convert HSI to RGB, which has the same spatial information. Given a HR HSI $\mathbf{H}^{HR}$, we utilize the spectral response matrix of Nikon D700 as HSIFN \citep{lai2024hyperspectral} to synthesize RGB $\mathbf{I}^{HR}$.

\begin{algorithm}[t]
	\caption{Synthetic generation pipeline}
	\label{algo:pipeline}
	\KwIn{Source-view HR HSIs $\mathbf{H}^{HR}$}
	\KwOut{Target-view HR RGBs $\mathbf{I}^{Ref}$, Ground-truth optical flows $\mathbf{I}^{Flow}$}
	\While{not enough source-view HSIs}{
		\LinesNumbered
		Convert selected source-view HSI $\mathbf{H}^{HR}$ to RGB $\mathbf{I}^{HR}$ via spectral response function (SRF)\;
		Estimate the corresponding depth map $\mathbf{D}^{HR}$\;
		Generate multiplane images (MPIs) by $\{(\mathbf{c}_n, \bm{\sigma}_n, \mathbf{d}_n)\}^N_{n=1} = \mathcal{F}(\mathbf{I}^{HR}, \mathbf{D}^{HR})$\;
		Set the number of target viewpoints $\mathbf{T}$\;
		\ForEach{number in $\mathbf{T}$}{
			Generate a random camera pose with viewpoint $\bm{\mu}_t$\;
			Render the target-view MPI\;
			Generate the target-view image $\mathbf{I}^{Ref}$\;
			Generate the ground-truth optical flow $\mathbf{I}^{Flow}$ between source-view image $\mathbf{I}^{HR}$ and target-view image $\mathbf{I}^{Ref}$\;
		}
	}
\end{algorithm}

\noindent{\textbf{Generation Stage.}} The key to the synthetic generation pipeline is that we need to produce ground-truth image pairs with accurate optical flow. The flow chart of the pipeline is shown in Algorithm~\ref{algo:pipeline}. Following the work \citep{liang_2023_ICCV}, we generate multiplane images (MPI) with camera viewpoints to obtain a new view image and its optical flow. To construct MPI, we first utilize a monocular depth estimation network \citep{ranftl2020towards} (MiDaS) to obtain its depth map $\mathbf{D}^{HR}$. 
An MPI is composed of $N$ front-parallel RGB$\sigma$ planes, as seen from the source camera perspective. Each plane is characterized by color channels $\mathbf{c}_n$, a density channel $\bm{\sigma}_n$, and a depth value $\mathbf{d}_n$. These attributes are derived from the high-resolution source image $\mathbf{I}^{HR}$ and its corresponding depth map $\mathbf{D}^{HR}$. The MPI can be obtained by
\begin{equation}
	\{(\mathbf{c}_n, \bm{\sigma}_n, \mathbf{d}_n)\}^N_{n=1} = \mathcal{F}(\mathbf{I}^{HR}, \mathbf{D}^{HR}),
\end{equation}
where $\mathcal{F}$ denotes the MPI generation network as in \citep{han2022single}, and $N$ is a predefined parameter. 
By employing a target camera viewpoint $\bm{\mu}_{t}$, we can render an image from the perspective of the target view using the MPI obtained from the source view. This process utilizes planar inverse warping and is fully differentiable.
Thus, each pixel $[x_t, y_t]$ on the novel image plane can be mapped to the pixel $[x_s, y_s]$ on the $n$-th plane of source-view MPI via homography function \citep{heyden2005multiple} as
\begin{equation}
	[x_s, y_s, 1] \sim \mathbf{K}_s ( \mathbf{R}-\frac{\mathbf{t}\mathbf{n}^{T}}{\mathbf{d}_n} ) \mathbf{K}^{-1}_{t} [x_t, y_t, 1]^{T},
	\label{eq: position}
\end{equation}
where $\mathbf{R}$ and $\mathbf{t}$ are the rotation and translation, $\mathbf{n}=[0,0,1]^{T}$ is the normal vector, $\mathbf{K}_s$ and $\mathbf{K}_t$ are the camera intrinsics. The target-view color $\mathbf{c}^{'}$ and density $\bm{\sigma}^{'}$ can be generated by bilinear sampling. Thus, the target-view image can be rendered by
\begin{equation}
	\mathbf{I}^{Ref} = \sum_{n=1}^{N}(\mathbf{c}^{'}_{n}\bm{\alpha}^{'}_{n} \prod_{m=1}^{n-1}(1-\bm{\alpha}_{m}^{'})),
	\label{eq:render}
\end{equation}
where $\bm{\alpha}_{m}^{'}=\text{exp}(-\bm{\delta}_{n}\bm{\sigma}_{n}^{'})$, and $\bm{\delta}_{n}$ denotes the distance between plane $n$ and $n+1$. 

Through the above process of synthesizing the target-view image, we simultaneously compute the optical flow on each plane at pixel $[x_s, y_s]$ of source image $\mathbf{I}^{HR}$ by $\mathbf{f}_n=[x_t-x_s, y_t-y_s]$ with a backward-warp process by Eq. (\ref{eq: position}). After obtaining optical flow $\mathbf{f}\in\mathbb{R}^{H\times W\times2}$ from each panel of MPI, to guarantee the matching between final optical flow with target-view image $\mathbf{I}^{Ref}$, we utilize the volume rendering as in Eq. (\ref{eq:render}) to render final optical flow $\mathbf{I}^{Flow}$ by
\begin{equation}
	\mathbf{I}^{Flow} = \sum_{n=1}^{N}(\mathbf{f}_{n}\bm{\alpha}_{n}\prod_{m=1}^{n-1}(1-\bm{\alpha}_{m})).
\end{equation}

\noindent{\textbf{Alignment Stage.}} We aim to align LR HSI and HR RGB reference, where we first generate LR HSI from HR HSI $\mathbf{H}^{HR}$ and convert it to LR RGB $\mathbf{I}^{HR}$ by SRF, thus it exists resolution gap (\ie, LR and HR) and distribution gap (\eg, brightness, contrast, saturation, hue, and occlusion). With the pair data obtained above, we need to preprocess the data to simulate the real resolution and distribution gaps that will exist. We first generate the preprocessed target image $\mathbf{I}^{HSI}$ as
\begin{equation}
	\mathbf{I}^{HSI} = Up(SRF(Down(Aug(\mathbf{H}^{HR})))),
\end{equation}
where $Down(\cdot)$ means spatial downsampling, $Up(\cdot)$ represents spatial upsampling, and $SRF(\cdot)$ is the spectral response function to convert HSI to RGB. Considering the resolution and distribution gaps in real applications, we perform data enhancement $Aug(\cdot)$, containing Gaussian noise, blur, and brightness augmentation, \etc 

In the \textbf{\textit{first stage}}, we aim to fine-tune the optical flow model to predict more accurate optical flow and coarsely warp the reference image. 
Building upon the research by \citep{teed2020raft}, we opt for the widely adopted pixel-wise model RAFT for estimating optical flow due to its proficiency in predicting pixel-level displacements.
The predicted optical flow $\mathbf{P}^{flow}$ aligned RGB image $\mathbf{I}^{Ref}_{coarse}$ is obtained as
\begin{align}
	\mathbf{P}^{flow} &= Estimator(\mathbf{I}^{Ref}, \mathbf{I}^{HSI}),\\
	\mathbf{I}^{Ref}_{coarse} &= Warp(\mathbf{I}^{Ref}, \mathbf{P}^{flow}),
\end{align}
where $Estimator(\cdot)$ denotes the optical flow model, $Warp(\cdot)$ represents the warp operation, $\mathbf{P}^{flow}$ is the predicted optical flow, and $\mathbf{I}^{Ref}_{coarse}$ is the coarsely warped reference image.

In the \textbf{\textit{second stage}}, we aim to train the warp model to refine the damaged textures and generate a more detailed and accurate image. Since the warping operation warps the image according to the predicted optical flow, sampling from the neighborhood in some regions without corresponding optical flow inevitably produces some regions with wrong textures, such as the ghosting artifacts \citep{zhao2020maskflownet}.
Thus, we propose a simple network, WarpNet based on U-Net \citep{ronneberger2015u} to refine textures. The final warped reference image can be obtained by
\begin{equation}
	\mathbf{I}^{Ref}_{warp} = WarpNet(\mathbf{I}^{Ref}_{coarse}, \mathbf{I}^{HSI}).
\end{equation}

\subsection{Feature Extraction}
\label{sec: feature extraction}

\noindent{\textbf{Image Encoders.}}
Considering the resolution gap between LR HSI and HR RGB image, which affects the robust corresponding match and fusion. We aim to extract multi-scale features of $\mathbf{H}^{LR\uparrow}$ and $\mathbf{I}^{Ref}_{warp}$, where $\mathbf{H}^{LR\uparrow}$ is upsampling from $\mathbf{H}^{LR}$. For the HSI encoder, we acquire the multi-scale features by 
\begin{equation}
	\mathbf{F}^{HSI}_{l} = E^{HSI}_{l}(\mathbf{H}^{LR\uparrow}), \quad l=1,\ldots,L,
\end{equation}
where $E^{HSI}_{l}$ represents feature encoder of HSI at $l$-th scale. In our larger model, we set the layers of scale $L=3$. 
For RGB encoders, we utilize the same network architectures to capture the multi-scale features that can be expressed as 
\begin{align}
	\mathbf{F}^{RGB}_{l} &= E^{RGB}_{l}(\mathbf{I}^{HSI}),\\
    \mathbf{F}^{Ref1}_{l} &= E^{Ref1}_{l}(\mathbf{I}^{Ref}_{warp}),\\
    \mathbf{F}^{Ref2}_{l} &= E^{Ref2}_{l}(\mathbf{I}^{Ref}_{warp}),
\end{align}
where $E^{RGB}_{l}, E^{Ref1}_{l}, E^{Ref2}_{l}$ denote the same architecture encoders for generated target RGB $\mathbf{I}^{HSI}$, and $\mathbf{I}^{Ref}_{warp}$ at $l$-th scale. 

\subsection{Feature Aggregation}
\label{sec: feature aggregation}
Despite the TSIA module being able to align and warp the reference RGB more accurately, it is still suffering from inaccurate alignment. Moreover, as a separate operating at the image level, it lacks efficiency in extracting feature-level information critical for reconstruction and effective interaction with the fusion stage. 
Therefore, to further learn more efficient and accurate spatial offsets and feature matching by long-range similar information, and to enhance the interaction of fine-grained feature aggregation and fusion modules, we propose a Feature Aggregation module shown in Fig.~\ref{fig:framework}. 
We utilize global attention matching to obtain global similarity and position maps, which are priors for further iterative deformable feature aggregation blocks to dynamically aggregate the related features. 
In addition, based on the fusion results of each layer, IDFA can generate the corresponding learnable offsets iteratively.
Benefiting from the interaction between modules, we can adjust promptly based on the fusion results, enabling the aggregation of more accurate and effective features.

\noindent{\textbf{Global Attention Matching (GAM).}} 
We introduce GAM to obtain similarity relationships between image pairs. Following the works \citep{jiang2021robust, cao2022reference}, we use a pre-trained contrastive correspondence network, \ie, two feature encoders $E^{RGB}_{l}$ and $E^{Ref1}_{l}$ with the same architecture for feature extracting. These encoders are trained with contrastive learning on the RGB dataset. Due to the RGB prior knowledge, we can estimate the stable correspondence between $\mathbf{I}^{HSI}$ and $\mathbf{I}^{Ref}_{warp}$.

Then, we divide the extracted features $\mathbf{F}^{RGB}_{l}$ and $\mathbf{F}^{Ref1}_{l}$ to patches $\mathbf{P}_{HSI}$ and $\mathbf{P}_{RGB}$, and calculate the similarity between LR patch $i$ and HR patch $j$. The purpose is to find the most relevant patch, so we obtain correlation matrix $\mathbf{M}_{i,j}$, similarity map $\mathbf{S}_{i}$, and corresponding position map $\mathbf{C}_{i}$ by
\begin{align}
	&\mathbf{M}_{i,j} = norm(\{{\mathbf{P}_{HSI}}^{T}\}_{i}\cdot\{\mathbf{P}_{RGB}\}_{j}), \\
	&\mathbf{S}_{i} = \max_{j}\mathbf{M}_{i,j}, \quad\mathbf{C}_{i} = \arg\max_{j}\mathbf{M}_{i,j},
\end{align}
where $\{{\mathbf{P}_{HSI}}^{T}\}_{i}$ and $\{\mathbf{P}_{RGB}\}_{j}$ are the $i$-th and $j$-th patch of $\mathbf{P}_{HSI}$ and $\mathbf{P}_{RGB}$ from extracted RGB features $\mathbf{F}^{RGB}_{l}$ and $\mathbf{F}^{Ref1}_{l}$, respectively, $norm(\cdot)$ means the normalization.

\noindent{\textbf{Iterative Deformable Feature Aggregation (IDFA).}} 
To enhance the interaction between alignment and fusion modules and aggregate global similar features for HSI fusion, we propose IDFA based on the modified DCN \citep{dai2017deformable, zhu2019deformable} (MDCN). 
Suppose a position $p$ in HR RGB and the correspondence position $p^\prime=\mathbf{C}_{p}$ from the correspondence position map $\mathbf{C}$. We can acquire the spatial difference $\Delta{p}=p^\prime-p$ and the aggregated feature $\mathbf{A}^{Ref}_{l}$ at position $p$ as follows,
\begin{equation}
\begin{aligned}
	\mathbf{A}^{Ref}_{l}(p) &= \mathcal{D}(\mathbf{F}^{Ref2}_{l}) \\
	&= \sum^{K}_{k=1}\omega_{k}\cdot\mathbf{F}^{Ref2}_{l}(p^\prime+p_{k}+\Delta{p}_{k})\cdot m_{k},
\end{aligned}
\end{equation}
where $p_{k}\in\{(-1,-1),(-1,0),\ldots,(1,1)\}$, $\omega_{k}$ denotes the convolutional kernel weight, and $\mathcal{D}$ means the deformable convolution. $\Delta{p}_{k}$ is the learnable offset and $m_{k}$ is the learnable modulation mask, \ie,
\begin{align}
	&\Delta{p}_{k}=r\cdot \text{Tanh}(Conv([\mathbf{F}^{SR}_{l-1};\mathbf{F}^{Ref2}_{l}])),\\
	&m_{k}=\text{Sigmoid}(s_{k}\cdot Conv([\mathbf{F}^{SR}_{l-1};\mathbf{F}^{Ref2}_{l}])),
\end{align}
where $\text{Tanh}$ and $\text{Sigmoid}$ denote activation function in our network, $Conv(\cdot)$ and $[;]$ represent convolutional layer and concatenation operation, respectively. 
$r$ denotes the max magnitude, which limits the maximum change in learnable spatial offsets.
$s_{k}$ is obtained from similarity map $\mathbf{S}$ and used as pre-similarity to restrict the result of the convolutional layer, which can help the model to alleviate the influence of matching inaccurate features. 
As shown in Fig.~\ref{fig:framework}(a) and Fig.~\ref{fig:framework}(b), for the first scale of feature aggregation module, \ie, when $l=1$, the input lacks the fusion feature $\mathbf{F}^{SR}_{l-1}$ and the concatenation operation.

\subsection{Attention Fusion}
\label{sec: attention fusion}
To explore the spectral correspondence, and capture long-range similarity and dependencies along spectra for spectral matching, we design the attention fusion module shown in Fig.~\ref{fig:modules}. The attention fusion module consists of two basic blocks, \ie, Spectral-wise Self-Attention (SSA) and Reference-guided Cross-Attention (RCA) to accurately fuse the content of HSI features $\mathbf{F}^{HSI}_{l}$ and RGB reference features $\mathbf{A}^{Ref}_{l}$.
Besides, the calculation along spectral dimension can reduce the computational burden compared to global spatial attention and circumvents the high spatial sparsity of HSIs \citep{cai2022mask, cai2022coarse}.

\noindent{\textbf{Attention Block.}} 
We introduce the SSA to capture spectral long-range similarity from HSI features and the RCA to explore the spectral correspondence, \ie, learn the most relevant spectral features in reference RGB features $\mathbf{A}^{Ref}_{l}$ inspired by \citep{zamir2022restormer}. We obtain the fused features $\mathbf{F}^{SR}_{l}$ at $l$-th layer of scale by
\begin{equation}
	\mathbf{F}^{SR}_{l} = \{RCA(SSA(Conv([\mathbf{F}^{HSI}_{l};\mathbf{F}^{SR}_{l-1}])),\mathbf{A}^{Ref}_{l})\}_{N},
\end{equation}
where SSA and RCA denote the attention blocks, and $N$ represents the number of combined attention blocks. We set $N=4$ when $l=1$, $N=3$ when $l=2$, and $N=2$ when $l=3$.
As shown in Fig.~\ref{fig:framework}(a) and Fig.~\ref{fig:framework}(c), for the first scale of attention fusion module, \ie, when $l=1$, the input lacks the fusion feature $\mathbf{F}^{SR}_{l-1}$ and the concatenation operation.

Suppose the $\mathbf{X}$ denotes $Conv([\mathbf{F}^{HSI}_{l};\mathbf{F}^{SR}_{l-1}])$ or the output $\mathbf{X}_{out}$ of last block, and $\mathbf{Y}$ denotes $\mathbf{A}^{Ref}_{l}$, where $\mathbf{X}\in\mathbb{R}^{\hat{H}\times \hat{W}\times C}$ and $\mathbf{Y}\in\mathbb{R}^{\hat{H}\times \hat{W}\times C}$. $C$ represents the channel dimensions. SSA is similar to the RCA block, so we briefly introduce the cross-attention block here. We first generate $\mathbf{Q}=\mathbf{X}\mathbf{W}^{Q}$, $\mathbf{K}=\mathbf{X}\mathbf{W}^{K}$, and $\mathbf{V}=\mathbf{X}\mathbf{W}^{V}$ from the layer normalization, $1\times 1$ convolutions, and $3\times 3$ depth-wise convolutions. Then, the reference-guided Multi-head Cross-Attention (R-MCA) is shown as
\begin{equation}
    \begin{aligned}
        &\hat{\mathbf{X}}=\rm{Attention}(\hat{\mathbf{Q}}, \hat{\mathbf{K}},\hat{\mathbf{V}})\mathbf{W} + \mathbf{X}, \\
        &\rm{Attention}(\hat{\mathbf{Q}}, \hat{\mathbf{K}},\hat{\mathbf{V}})=\rm{Concat}_{r=1}^{R}(head_{r}),
    \end{aligned}
\end{equation}
where $\hat{\mathbf{Q}}$, $\hat{\mathbf{K}}$, and $\hat{\mathbf{V}}\in\mathbb{R}^{\hat{H}\hat{W}\times C}$ are reshaped from original tensors, and $head_{r}$ is obtained by
\begin{equation}
    \mathbf{A}_{r} = \rm{Softmax}(\hat{\mathbf{K}}_{r}^{\rm{T}}\hat{\mathbf{Q}}_{r}), \quad head_{r}=(\hat{\mathbf{V}}_{r}\odot\hat{\mathbf{Y}}_{r})\mathbf{A}_{r},
\end{equation}
where $r$ denotes that we divide the original tensor by $R$ parts, \ie, we have total $R$ heads. $\rm{T}$ represents the transposed operation, and $\mathbf{A}_{r}$ is the spectral attention map. $\odot$ denotes the element-wise multiplication, $\hat{\mathbf{Y}}_{r}$ is obtained from the feature aggregation module.
In contrast to RCA, SSA does not utilize $\hat{\mathbf{Y}}_{r}$ for the multiplication within each head. Instead, it primarily exploits global self-similarity.
Last, we use the GDFN \citep{zamir2022restormer}, including Norm and FFN layers shown in Fig.~\ref{fig:modules} from $\hat{\mathbf{X}}$ to generate $\mathbf{X}_{out}$ as input $\mathbf{X}$ for the next attention block, which can perform features transformation, and propagate the useful information further.

Based on the fused feature $\mathbf{F}^{SR}_{L}$ of the last layer and input, the upsampled HSI $\mathbf{H}^{LR\uparrow}$, we use the residual connection to generate the ultimate HR HSI shown as follows
\begin{equation}
	\mathbf{H}^{SR} = \mathbf{H}^{LR\uparrow}+ \mathbf{F}^{SR}_{L}.
\end{equation}

\begin{table*}[htb]
\begin{center}
\small
\setlength{\tabcolsep}{0.11cm}
\renewcommand\arraystretch{0.5}
\caption{Quantitative comparison of various methods under several scale factors on the simulated natural dataset,  \textit{Flower}. PSNR is in dB}
\label{tab:result on simulated dataset}
\begin{tabular}{ccccccccccccc}
\toprule
\rowcolor{graycolor}

\makecell{Scale \\ Factor} & \makecell{Metric}                        & \makecell{Bicubic}    & \makecell{BiQRNN \\ (JSTAR'21)}  & \makecell{SSPSR \\ (TCI'20)} & \makecell{MCNet \\ (RS'20)}  & \makecell{ESSA \\ (ICCV'23)}   & \makecell{Optimized \\ (CVPR'19)} & \makecell{Integrated \\ (TGRS'19)}  & \makecell{NonReg \\ (TGRS'22)} & \makecell{$u^2$-MDN \\ (TGRS'22)} & \makecell{HSIFN \\ (TNNLS'24)} &  \makecell{Ours}      \\ \midrule

\multirow{3}{*}{4}
	& PSNR$\uparrow$ & 29.88 & 37.53 & 37.02 & 37.35 & 38.08 & 25.40 & 29.09 & 25.92 & 25.85 & 41.76 & \textbf{42.73} \\ & \\
    & SSIM$\uparrow$ & 0.914 & 0.979 & 0.976 & 0.979 & 0.977 & 0.848 & 0.908 & 0.838  & 0.804 & 0.990 & \textbf{0.992} \\ & \\
    & SAM$\downarrow$  & 0.055 & 0.030 & 0.045 & 0.034 & 0.034 & 0.319 & 0.232 & 0.311  & 0.127 & 0.026 & \textbf{0.023} \\ \midrule
\multirow{3}{*}{8}    
	& PSNR$\uparrow$ & 25.49 & 30.12 & 30.22 & 30.03 & 31.17 & 25.42 & 25.97 & 25.56 & 25.39 & 37.69 & \textbf{38.15} \\ & \\
    & SSIM$\uparrow$ & 0.835 & 0.921 & 0.925 & 0.921 & 0.921 & 0.963 & 0.869 & 0.823  & 0.834 & 0.976 & \textbf{0.982} \\ & \\
    & SAM$\downarrow$  & 0.094 & 0.057 & 0.057 & 0.057 & 0.066 & 0.318 & 0.312 & 0.323  & 0.150 & 0.037 & \textbf{0.035} \\ \midrule
\multirow{3}{*}{16}    
	& PSNR$\uparrow$ & 22.10 & 25.34 & 25.31 & 25.42 & 25.85 & 25.39 & 23.57 & 25.54 & 25.32 & 32.82 & \textbf{34.07} \\ & \\
    & SSIM$\uparrow$ & 0.764 & 0.846 & 0.847 & 0.848 & 0.838 & 0.847 & 0.820 & 0.763  & 0.832 & 0.960 & \textbf{0.965} \\ & \\
    & SAM$\downarrow$  & 0.151 & 0.100 & 0.104 & 0.106 & 0.107 & 0.318 & 0.370 & 0.360  & 0.150 & 0.062 & \textbf{0.055} \\
\bottomrule
\end{tabular}
\end{center}
\end{table*}

\subsection{Learning Details}

Considering the aim to generate the spatially aligned reference RGB image and high-quality HR HSI, we introduce two phases for training. In the first phase, we train our model to align reference images guided by target RGB generated from LR HSI. We utilize an optical flow model to predict the flow between RGB images and WarpNet to refine the image texture. For the optical flow model training, we follow RAFT \citep{teed2020raft} to supervise the network between predicted flow, $\mathbf{f}_{i}$, with ground-truth flow, $\mathbf{f}_{gt}$, on the constraint of the $L_1$ distance over all the predicted sequence, ${\mathbf{f}_1,\ldots,\mathbf{f}_N}$, with exponentially increasing weights. Thus, the loss can be obtained by
\begin{equation}
	\mathcal{L}_{flow} = \sum_{i=1}^{N}\gamma^{N-i}\|\mathbf{f}_{gt}-\mathbf{f}_{i}\|_{1}.
\end{equation}
To supervise the WarpNet model, we then use $L_1$ loss to constrain learning to generate more accurately aligned RGB images. Similarly, in the second phase, we also utilize $L_1$ loss followed in HSIFN \citep{lai2024hyperspectral} to constrain the network to generate HR HSI.

\begin{figure}
\begin{center}
\setlength{\tabcolsep}{0.02cm}
    \begin{subfigure}[h]{0.15\textwidth}
          \centering
          \small
          \includegraphics[width=1\linewidth]{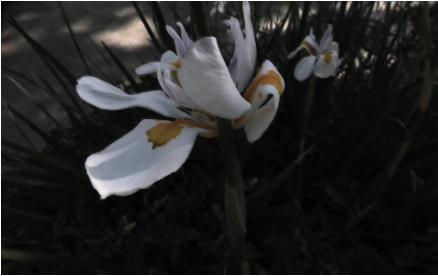}
          \caption{Target on Flower}
      \end{subfigure}
    \hfill
    \begin{subfigure}[h]{0.15\textwidth}
          \centering
          \small
          \includegraphics[width=1\linewidth]{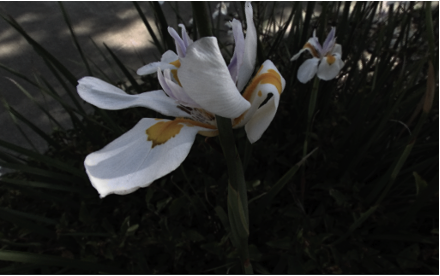}
          \caption{Reference}
      \end{subfigure}
    \hfill
    \begin{subfigure}[h]{0.15\textwidth}
          \centering
          \small
          \includegraphics[width=1\linewidth]{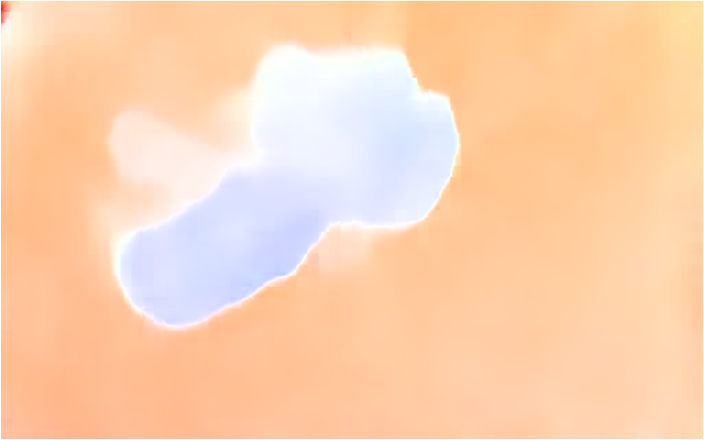}
          \caption{Displacement}
      \end{subfigure}
    \vfill
    \begin{subfigure}[h]{0.15\textwidth}
          \centering
          \small
          \includegraphics[width=1\linewidth]{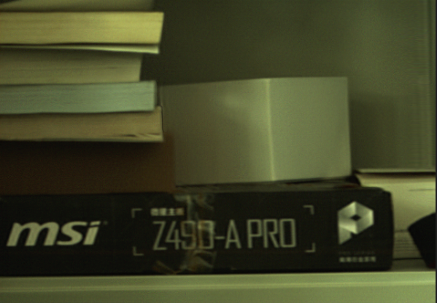}
          \caption{Target on Real}
      \end{subfigure}
    \hfill
    \begin{subfigure}[h]{0.15\textwidth}
          \centering
          \small
          \includegraphics[width=1\linewidth]{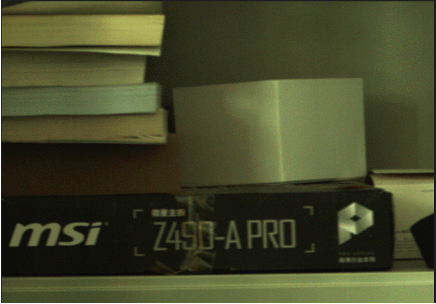}
          \caption{Reference}
      \end{subfigure}
    \hfill
    \begin{subfigure}[h]{0.15\textwidth}
          \centering
          \small
          \includegraphics[width=1\linewidth]{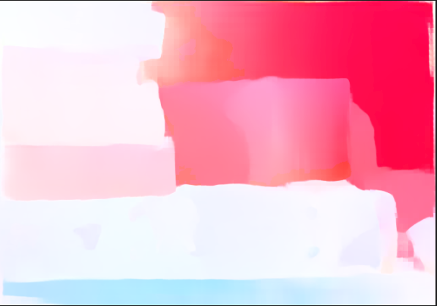}
          \caption{Displacement}
      \end{subfigure}
      \vfill
    \begin{subfigure}[h]{0.15\textwidth}
          \centering
          \small
          \includegraphics[width=1\linewidth]{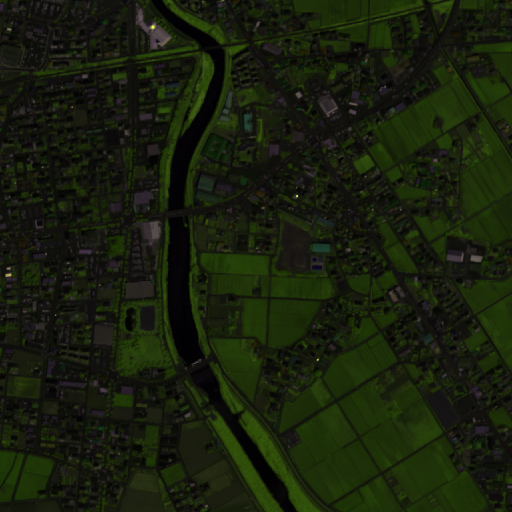}
          \caption{Target on Chikusei}
      \end{subfigure}
    \hfill
    \begin{subfigure}[h]{0.15\textwidth}
          \centering
          \small
          \includegraphics[width=1\linewidth]{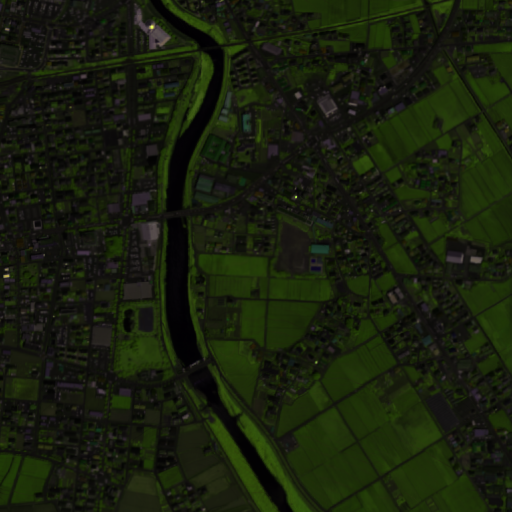}
          \caption{Reference}
      \end{subfigure}
    \hfill
    \begin{subfigure}[h]{0.15\textwidth}
          \centering
          \small
          \includegraphics[width=1\linewidth]{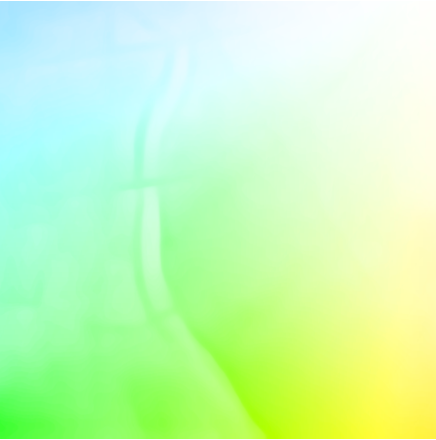}
          \caption{Displacement}
      \end{subfigure}
\end{center}
   \caption{Example scenes with displacement in three datasets. Every scene has a target HSI, an unaligned reference RGB, and the displacement between target and reference image. From top to bottom are the Flower dataset, Real dataset, and Chikusei dataset. Here we convert HSI to RGB for better visualization}
    \label{fig:dataset}
\end{figure}

\begin{figure*}
 \centering
\small
\hspace{1mm}
 \setlength{\tabcolsep}{0.1cm}
\begin{minipage}[l]{0.22\linewidth}
\begin{flushleft}
 \vspace{-0.1mm}
  \hspace{-3mm}
\includegraphics[width=1.045\linewidth]{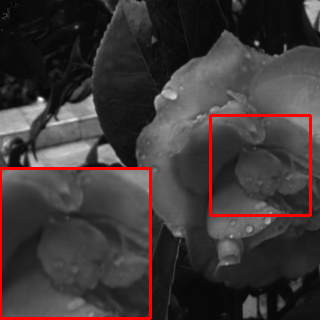}\\
   \begin{tabular}[c]{@{}c@{}}\hspace{7mm}Ground Truth\end{tabular}
  \end{flushleft}
 \end{minipage}
 \hspace{-1mm}
 \begin{minipage}[t]{0.5\linewidth}
  \begin{tabular}{cccccc}
   \includegraphics[width=0.2\linewidth]{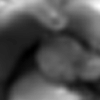}
  & \includegraphics[width=0.2\linewidth]{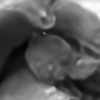}
  & \includegraphics[width=0.2\linewidth]{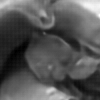}
  & \includegraphics[width=0.2\linewidth]{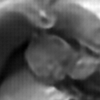}
  & \includegraphics[width=0.2\linewidth]{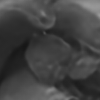}
  & \includegraphics[width=0.2\linewidth]{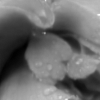}
   \\
    Input ($\times 4$) & BiQRNN & SSPSR  & MCNet & ESSA  & Optimized \\
   
 \includegraphics[width=0.2\linewidth]{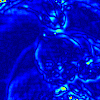}
  &\includegraphics[width=0.2\linewidth]{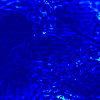}
  &\includegraphics[width=0.2\linewidth]{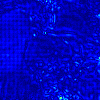}
  &\includegraphics[width=0.2\linewidth]{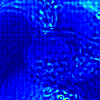}
  &\includegraphics[width=0.2\linewidth]{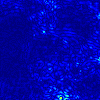}
  &\includegraphics[width=0.2\linewidth]{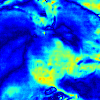}
	  
  \\
   Error Map &  BiQRNN & SSPSR  & MCNet & ESSA  & Optimized  
  \end{tabular}
 \end{minipage}
 \hspace{22mm}
 \begin{minipage}[l]{0.1\linewidth}
 \begin{flushright}
    \vspace{-5mm}
     \includegraphics[width=0.54\linewidth]{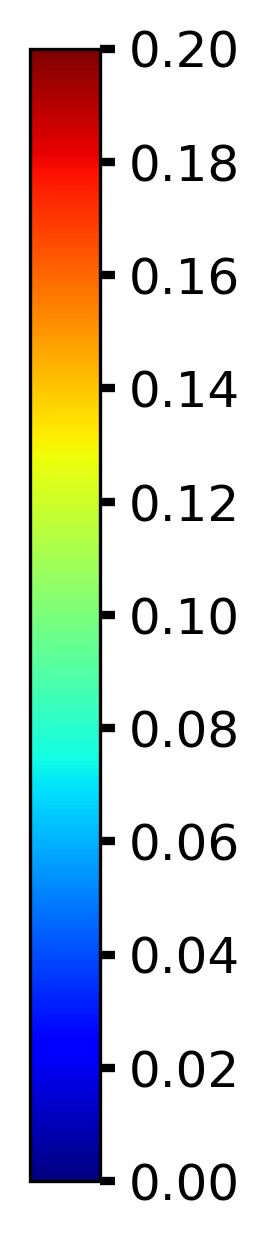}
 \end{flushright}
 \end{minipage}
 
 \hspace{42mm}
 \begin{minipage}[t]{0.5\linewidth}
  \begin{tabular}{cccccc}
   
  \includegraphics[width=0.2\linewidth]{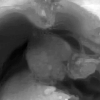}
  & \includegraphics[width=0.2\linewidth]{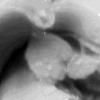}
  & \includegraphics[width=0.2\linewidth]{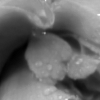}
  & \includegraphics[width=0.2\linewidth]{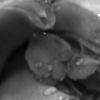}
  & \includegraphics[width=0.2\linewidth]{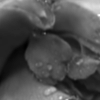}
  & \includegraphics[width=0.2\linewidth]{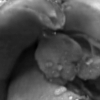}
   \\
    Integrated & NonReg & $u^2$-MDN  & HSIFN & Ours  & GT \\

  \includegraphics[width=0.2\linewidth]{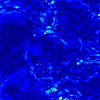}
  &\includegraphics[width=0.2\linewidth]{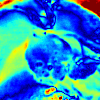}
  &\includegraphics[width=0.2\linewidth]{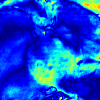}
  &\includegraphics[width=0.2\linewidth]{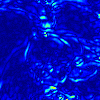}
  &\includegraphics[width=0.2\linewidth]{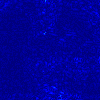}
  &\includegraphics[width=0.2\linewidth]{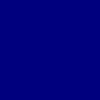}
	  
  \\
   Integrated & NonReg & $u^2$-MDN  & HSIFN & Ours  & GT  
  \end{tabular}
 \end{minipage}
 \hspace{22mm}
 \begin{minipage}[l]{0.1\linewidth}
 \begin{flushright}
    \vspace{-5mm}
     \includegraphics[width=0.54\linewidth]{figure_color_bar.png}
 \end{flushright}
 \end{minipage}

\hspace{2mm}
 \begin{minipage}[l]{0.22\linewidth}
\begin{flushleft}
 \vspace{-0.1mm}
  \hspace{-3mm}
\includegraphics[width=1.045\linewidth]{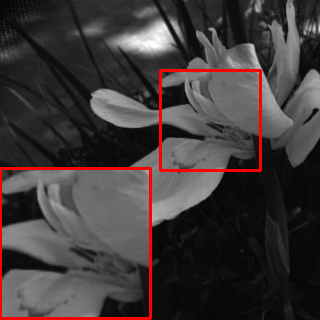}\\
   \begin{tabular}[c]{@{}c@{}}\hspace{7mm}Ground Truth\end{tabular}
  \end{flushleft}
 \end{minipage}
 \hspace{-1mm}
 \begin{minipage}[t]{0.5\linewidth}
  \begin{tabular}{cccccc}
   \includegraphics[width=0.2\linewidth]{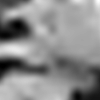}
  & \includegraphics[width=0.2\linewidth]{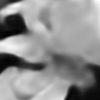}
  & \includegraphics[width=0.2\linewidth]{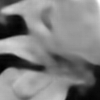}
  & \includegraphics[width=0.2\linewidth]{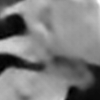}
  & \includegraphics[width=0.2\linewidth]{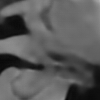}
  & \includegraphics[width=0.2\linewidth]{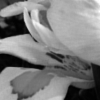}
   \\
    Input ($\times 8$) & BiQRNN & SSPSR  & MCNet & ESSA  & Optimized \\
   
 \includegraphics[width=0.2\linewidth]{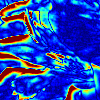}
  &\includegraphics[width=0.2\linewidth]{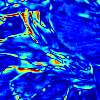}
  &\includegraphics[width=0.2\linewidth]{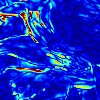}
  &\includegraphics[width=0.2\linewidth]{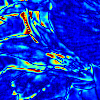}
  &\includegraphics[width=0.2\linewidth]{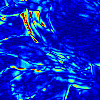}
  &\includegraphics[width=0.2\linewidth]{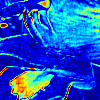}
	  
  \\
   Error Map &  BiQRNN & SSPSR  & MCNet & ESSA  & Optimized  
  \end{tabular} \end{minipage}
 \hspace{22mm}
 \begin{minipage}[l]{0.1\linewidth}
 \begin{flushright}
    \vspace{-5mm}
     \includegraphics[width=0.54\linewidth]{figure_color_bar.png}
 \end{flushright}
 \end{minipage}
 
 \hspace{42mm}
 \begin{minipage}[t]{0.5\linewidth}
  \begin{tabular}{cccccc}
   
  \includegraphics[width=0.2\linewidth]{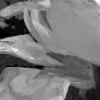}
  & \includegraphics[width=0.2\linewidth]{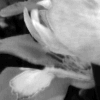}
  & \includegraphics[width=0.2\linewidth]{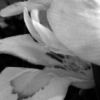}
  & \includegraphics[width=0.2\linewidth]{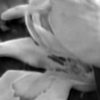}
  & \includegraphics[width=0.2\linewidth]{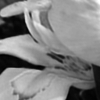}
  & \includegraphics[width=0.2\linewidth]{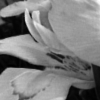}
   \\
    Integrated & NonReg & $u^2$-MDN  & HSIFN & Ours  & GT \\

  \includegraphics[width=0.2\linewidth]{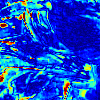}
  &\includegraphics[width=0.2\linewidth]{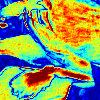}
  &\includegraphics[width=0.2\linewidth]{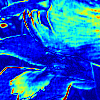}
  &\includegraphics[width=0.2\linewidth]{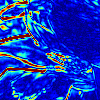}
  &\includegraphics[width=0.2\linewidth]{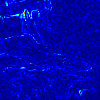}
  &\includegraphics[width=0.2\linewidth]{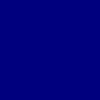}
	  
  \\
   Integrated & NonReg & $u^2$-MDN  & HSIFN & Ours  & GT \\  
  \end{tabular}
 \end{minipage}
 \hspace{22mm}
 \begin{minipage}[l]{0.1\linewidth}
 \begin{flushright}
    \vspace{-5mm}
     \includegraphics[width=0.54\linewidth]{figure_color_bar.png}
 \end{flushright}
 \end{minipage}

 \hspace{2mm}
 \begin{minipage}[l]{0.22\linewidth}
\begin{flushleft}
 \vspace{-0.1mm}
  \hspace{-3mm}
\includegraphics[width=1.045\linewidth]{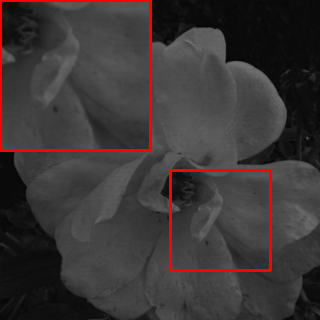}\\
   \begin{tabular}[c]{@{}c@{}}\hspace{7mm}Ground Truth\end{tabular}
  \end{flushleft}
 \end{minipage}
 \hspace{-1mm}
 \begin{minipage}[t]{0.5\linewidth}
  \begin{tabular}{cccccc}
   \includegraphics[width=0.2\linewidth]{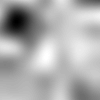}
  & \includegraphics[width=0.2\linewidth]{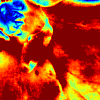}
  & \includegraphics[width=0.2\linewidth]{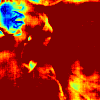}
  & \includegraphics[width=0.2\linewidth]{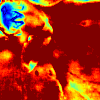}
  & \includegraphics[width=0.2\linewidth]{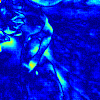}
  & \includegraphics[width=0.2\linewidth]{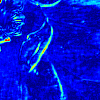}
   \\
    Input ($\times 16$) & BiQRNN & SSPSR  & MCNet & ESSA  & Optimized \\
   
 \includegraphics[width=0.2\linewidth]{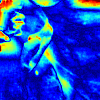}
  &\includegraphics[width=0.2\linewidth]{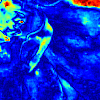}
  &\includegraphics[width=0.2\linewidth]{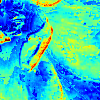}
  &\includegraphics[width=0.2\linewidth]{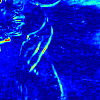}
  &\includegraphics[width=0.2\linewidth]{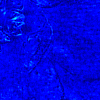}
  &\includegraphics[width=0.2\linewidth]{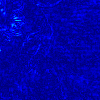}
  
  \\
   Error Map & Integrated & NonReg & $u^2$-MDN  & HSIFN &  Ours
  \end{tabular}
 \end{minipage}
 \hspace{22mm}
 \begin{minipage}[l]{0.1\linewidth}
 \begin{flushright}
    \vspace{-5mm}
     \includegraphics[width=0.54\linewidth]{figure_color_bar.png}
 \end{flushright}
 \end{minipage}

 \caption{Visual comparison on the Flower dataset. The ground truth and input HSIs are shown with band 20. The results of different methods under scale factor $\times 4$, $\times 8$, $\times 16$ from top to bottom with the error maps. Zoom in for details}
 \label{fig:simulated visual results}
\end{figure*}

\section{Experiments}
\label{sec:experiments}

In this section, we first introduce the datasets and metrics used for quantitative evaluation in our experiments. Subsequently, our approach is compared with several state-of-the-art methods on simulated and real data. Additionally, we conduct ablation studies to assess the effectiveness of different modules and model sizes.

\subsection{Experimental Settings}

\noindent{\textbf{Datasets.}} We evaluate the proposed method on three hyperspectral datasets, including a simulated natural flower dataset, a real natural dataset, and a remote-sensing dataset (\ie, Chikusei).

\noindent{\textit{\textbf{Simulated Natural Flower Dataset.}}}
Following the work \citep{lai2024hyperspectral}, we employ a simulated flower dataset with slight misalignment for our experiments. The HSIs with 31 bands in the simulated natural dataset are synthesized from real unaligned RGB-RGB image pairs in the light field dataset Flowers \citep{srinivasan2017learning}, using HSCNN+ \citep{shi2018hscnn+} for HSI recovery from RGB images.
It contains 3343 pairs of images with a resolution of $320\times520$. 100 pairs are chosen for testing and the rest pairs are for training. The simulated natural dataset will subsequently be referred to as "Flower".

\noindent{\textit{{\textbf{Real Natural Dataset.}}}
Following the work \citep{lai2024hyperspectral}, we also employ a real natural dataset, \ie, Real-HSI-Fusion \citep{lai2024hyperspectral}, which consists of 60 pairs of unaligned HR HSIs with 31 bands ranging from $400 nm$ to $700 nm$. The reference RGB is generated from one of the HSIs by utilizing the spectral response function of Nikon D700 same as \citep{akhtar2015bayesian, qu2022unsupervised}. 
Similarly, 10 pairs are chosen for testing and the rest pairs are for training. The real natural dataset will subsequently be referred to as "Real".

\noindent{\textit{\textbf{Remote-Sensing Dataset.}}}
We employ a remote-sensing dataset Chikusei \citep{yokoya2016airborne} following the works \citep{wu2023hsr, xie2023toward}. The dataset consists of 128 bands with a spectral range of $363 nm$ to $1018 nm$ and a spatial resolution of $2517\times2335$. The original HSI data is taken by a Headwall Hyperspec-VNIR-C imaging sensor over Chikusei, Japan. 
Due to the lack of detailed information on the edges, we select the center of Chikusei with $2048\times2048$ pixels and crop it to 16 non-overlapped images with a size of $512\times512$ for the experiments. Different from previous works, we directly select 4 parts for testing and the rest for training. We set the bands of 700-100-36 as R-G-B channels to generate HR RGB.

The details about these datasets are shown in Fig.~\ref{fig:dataset}. For all datasets, we randomly select one scene to show the target HSI, which is converted to RGB for display, the reference unaligned RGB, and the displacement between the target and reference images. The displacement is predicted by a pre-trained optical flow model, where the colors indicate the direction of objects in each scene.

\noindent{\textbf{Data Generation Details.}}
For the generation of paired simulation data and subsequent training in the alignment phase, we use the ICVL dataset \citep{arad2016sparse}. The ICVL dataset consists of 201 natural HSIs with a spatial resolution of $1300\times1392$. Similar to the Real dataset, the spectral range of HSI in the ICVL dataset is all from $400 nm$ to $700 nm$. We set the number of target-view images to 10 in our experiment, which means we have 10 pairs of data per scene.
Following the works \citep{liang_2023_ICCV, aleotti2021learning}, we adopt the same setting to build a virtual camera for generating novel view images within our pipeline. We assume a single image with a resolution of $W\times H$. The virtual camera is configured with a fixed intrinsic matrix $\mathbf{K}$, with focals $(f_x,f_y)=0.58(W,H)$ and an optical center at $(c_x,c_y)=0.5(W,H)$. The camera motion is defined by three translation parameters $t_x,t_y,t_z$, each ranging from -0.2 to 0.2. The camera rotation is determined by three Euler angles $a_x, a_y, a_z$, each within the range of [$-\frac{\pi}{90},\frac{\pi}{90}$]. Besides, to generate virtual images, we convert predicted depths into the range [1, 100].
Upon completing the data processing, we initially embark on training the optical flow model in the first phase as the work \citep{teed2020raft}. Subsequently, in the second phase, we devote 100 epochs to training the WarpNet. Finally, we conduct a comprehensive fine-tuning of the entire network on super-resolution data to optimize performance. 

\noindent{\textbf{Implementation Details.}}
Following the works \citep{fu2021bidirectional, lai2024hyperspectral}, the input LR HSIs are generated by blurring the HR HSIs with a Gaussian kernel ($\mu=8, \sigma=3$) and downsampling with a certain scale factor, \eg, $\times4$, $\times8$, and $\times16$.
For the RGB encoders, we utilize the same architectures as \citep{jiang2021robust}, \ie, the VGG19 \citep{simonyan2014very} networks with \textit{relu1\_1}, \textit{relu2\_1} and \textit{relu3\_1} layers to obtain multi-scale features.
Following the training strategy of \citep{lai2024hyperspectral}, we set the batch size to 1, utilize PyTorch \citep{paszke2019pytorch} to implement our network, and train our model on an NVIDIA GeForce RTX 3090. The AdamW \citep{loshchilov2017decoupled} optimizer with the weight decay rate of $5\times10^{-5}$ is used to minimize the $L_1$ loss function. The learning rate is set to $1\times10^{-5}$ on the three datasets. We train the networks on the Flower, Real, and Chikusei datasets for 100, 500, and 500 epochs, respectively.

\begin{table*}[htb]
\begin{center}
\small
\setlength{\tabcolsep}{0.11cm}
\renewcommand\arraystretch{0.5}
\caption{Quantitative comparison of various methods under several scale factors on the real natural dataset, \textit{Real}. PSNR is in dB}
\label{tab:result on real dataset}
\begin{tabular}{ccccccccccccc}
\toprule
\rowcolor{graycolor}

\makecell{Scale \\ Factor} & \makecell{Metric}                        & \makecell{Bicubic}    & \makecell{BiQRNN \\ (JSTAR'21)}  & \makecell{SSPSR \\ (TCI'20)} & \makecell{MCNet \\ (RS'20)}  & \makecell{ESSA \\ (ICCV'23)}   & \makecell{Optimized \\ (CVPR'19)} & \makecell{Integrated \\ (TGRS'19)}  & \makecell{NonReg \\ (TGRS'22)} & \makecell{$u^2$-MDN \\ (TGRS'22)} & \makecell{HSIFN \\ (TNNLS'24)} &  \makecell{Ours}      \\ \midrule
\multirow{3}{*}{4}
	& PSNR$\uparrow$ & 34.07 & 37.80 & 39.04 & 39.07 & 39.94 & 27.26 & 30.66 & 31.60 & 30.58 & 41.01 & \textbf{42.20} \\ & \\
    & SSIM$\uparrow$ & 0.941 & 0.969 & 0.976 & 0.974 & 0.982 & 0.916 & 0.935 & 0.951  & 0.936 & 0.986 & \textbf{0.989} \\ & \\
    & SAM$\downarrow$  & 0.042 & 0.044 & 0.040 & 0.038 & 0.040 & 0.189 & 0.157 & 0.081  & 0.087 & 0.036 & \textbf{0.031} \\ \midrule
\multirow{3}{*}{8}    
	& PSNR$\uparrow$ & 28.43 & 31.66 & 31.13 & 31.31 & 31.85 & 26.99 & 30.97 & 30.00 & 30.18 & 33.13 & \textbf{37.37} \\ & \\
    & SSIM$\uparrow$ & 0.870 & 0.908 & 0.906 & 0.904 & 0.913 & 0.909 & 0.932 & 0.923  & 0.936 & 0.946 & \textbf{0.974} \\ & \\
    & SAM$\downarrow$  & 0.058 & 0.060 & 0.069 & 0.063 & 0.064 & 0.195 & 0.138 & 0.167  & 0.096 & 0.061 & \textbf{0.042} \\ \midrule
\multirow{3}{*}{16}    
	& PSNR$\uparrow$ & 24.84 & 26.53 & 27.20 & 27.04 & 26.70 & 26.93 & 27.10 & 29.09 & 29.84 & 31.07 & \textbf{32.80} \\ & \\
    & SSIM$\uparrow$ & 0.815 & 0.832 & 0.851 & 0.845 & 0.918 & 0.912 & 0.885 & 0.904  & 0.932 & 0.939 & \textbf{0.951} \\ & \\
    & SAM$\downarrow$  & 0.086 & 0.111 & 0.077 & 0.088 & 0.111 & 0.220 & 0.206 & 0.584  & 0.109 & 0.100 & \textbf{0.059} \\
\bottomrule
\end{tabular}
\end{center}
\end{table*}

\noindent{\textbf{Evaluation Metrics.}}
We utilize three quantitative image quality metrics, \ie, peak signal-to-noise ratio (PSNR), structural similarity (SSIM), and spectral angle mapping (SAM) for evaluation. PSNR and SSIM are computed on the mean value of all bands, while SAM measures the similarity between the spectra. Higher PSNR and SSIM and lower SAM indicate better performance.

\subsection{Evaluation of HSI Super-Resolution Methods}
\label{sec: results}

\noindent{\textbf{Competing Methods.}}
We compare the proposed method with several state-of-the-art SISR methods, fusion-based methods, and unaligned reference-based methods. The latter two are collectively termed as RefSR. The SISR methods include BiQRNN \citep{fu2021bidirectional}, SSPSR \citep{jiang2020learning}, MCNet \citep{li2020mixed}, and ESSAformer \citep{zhang2023essaformer}. For RefSR methods, Optimized \citep{fu2019hyperspectral}, Integrated \citep{zhou2019integrated}, NonReg \citep{zheng2022nonregsrnet}, $u^2$-MDN \citep{qu2022unsupervised}, and HSIFN \citep{lai2024hyperspectral} are included. All experiments are conducted using scale factors of $\times4$, $\times8$, and $\times16$ between the LR HSI and HR RGB.

\noindent{\textbf{Results on the Flower Dataset.}}
We provide both quantitative comparisons of various methods on the Flower dataset in Table~\ref{tab:result on simulated dataset} and qualitative comparisons in Fig.~\ref{fig:simulated visual results}. For quantitative comparison, our proposed method achieves the best PSNR, SSIM, and SAM and significantly outperforms compared SISR and RefSR methods under different scale factors of $\times4$, $\times8$, and $\times16$. 
As shown in \Tref{tab:result on simulated dataset}, most of the methods show good performance at $\times4$, but there is an obvious drop off at $\times8$ and $\times16$ high scale factors. Especially, the SISR methods have close to a 5 dB PSNR decrease at increasing ratios, which is mainly because the SISR method speculates about missing or ambiguous information based on the known information of the input LR HSI. At the high scale ratios, it is difficult for SISR methods to recover high-quality details due to low-quality input images. 
In comparison to other RefSR methods, our proposed method demonstrates superior performance. 
We have retrained HSIFN \citep{lai2024hyperspectral} under the setting same as ours without extra training strategies.
It can be seen in Fig.~\ref{fig:simulated visual results}, that these RefSR methods have a higher intensity in error maps, which demonstrates they have not properly aligned reference images.
For qualitative evaluation, the visual HSI results with $20$-th band for three scenes under different scale factors are shown in Fig.~\ref{fig:simulated visual results}. The error map shows the absolute difference between the SR result and ground truth. As can be observed, our method yields the smallest error and closely approximates the HR ground truth.

\begin{figure*}
 \centering
\small
\hspace{1mm}
 \setlength{\tabcolsep}{0.1cm}
\begin{minipage}[l]{0.22\linewidth}
\begin{flushleft}
 \vspace{-0.1mm}
  \hspace{-3mm}
\includegraphics[width=1.045\linewidth]{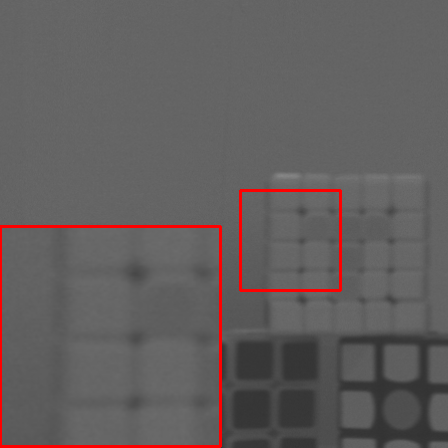}\\
   \begin{tabular}[c]{@{}c@{}}\hspace{7mm}Ground Truth\end{tabular}
  \end{flushleft}
 \end{minipage}
 \hspace{-1mm}
 \begin{minipage}[t]{0.5\linewidth}
  \begin{tabular}{cccccc}
   \includegraphics[width=0.2\linewidth]{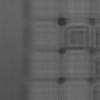}
  & \includegraphics[width=0.2\linewidth]{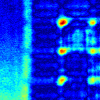}
  & \includegraphics[width=0.2\linewidth]{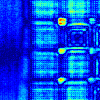}
  & \includegraphics[width=0.2\linewidth]{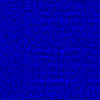}
  & \includegraphics[width=0.2\linewidth]{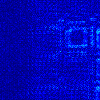}
  & \includegraphics[width=0.2\linewidth]{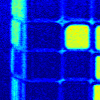}
   \\
    Input ($\times 4$) & BiQRNN & SSPSR  & MCNet & ESSA  & Optimized \\
   
 \includegraphics[width=0.2\linewidth]{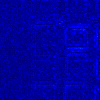}
  &\includegraphics[width=0.2\linewidth]{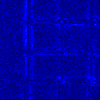}
  &\includegraphics[width=0.2\linewidth]{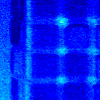}
  &\includegraphics[width=0.2\linewidth]{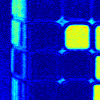}
  &\includegraphics[width=0.2\linewidth]{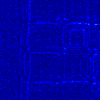}
  &\includegraphics[width=0.2\linewidth]{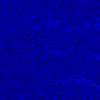}
  
  \\      
   Error Map & Integrated & NonReg & $u^2$-MDN  & HSIFN &  Ours
  \end{tabular}
 \end{minipage}
 \hspace{22mm}
 \begin{minipage}[l]{0.1\linewidth}
 \begin{flushright}
    \vspace{-5mm}
     \includegraphics[width=0.54\linewidth]{figure_color_bar.png}
 \end{flushright}
 \end{minipage}

\hspace{2mm}
 \begin{minipage}[l]{0.22\linewidth}
\begin{flushleft}
 \vspace{-0.1mm}
  \hspace{-3mm}
\includegraphics[width=1.045\linewidth]{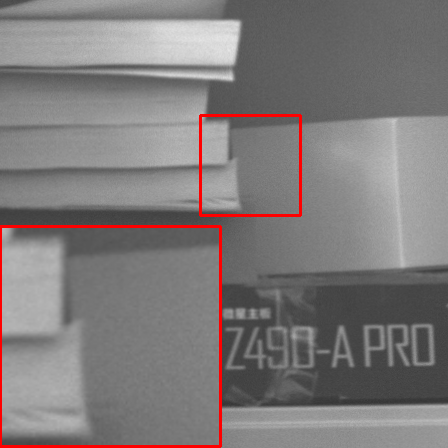}\\
   \begin{tabular}[c]{@{}c@{}}\hspace{7mm}Ground Truth\end{tabular}
  \end{flushleft}
 \end{minipage}
 \hspace{-1mm}
 \begin{minipage}[t]{0.5\linewidth}
  \begin{tabular}{cccccc}
   \includegraphics[width=0.2\linewidth]{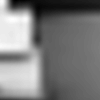}
  & \includegraphics[width=0.2\linewidth]{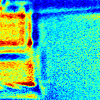}
  & \includegraphics[width=0.2\linewidth]{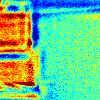}
  & \includegraphics[width=0.2\linewidth]{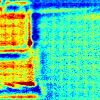}
  & \includegraphics[width=0.2\linewidth]{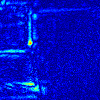}
  & \includegraphics[width=0.2\linewidth]{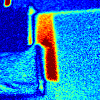}
   \\
    Input ($\times 8$) & BiQRNN & SSPSR  & MCNet & ESSA  & Optimized \\
   
 \includegraphics[width=0.2\linewidth]{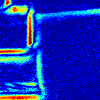}
  &\includegraphics[width=0.2\linewidth]{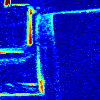}
  &\includegraphics[width=0.2\linewidth]{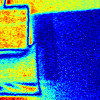}
  &\includegraphics[width=0.2\linewidth]{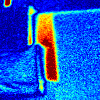}
  &\includegraphics[width=0.2\linewidth]{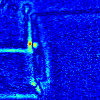}
  &\includegraphics[width=0.2\linewidth]{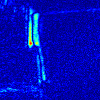}
  
  \\
   Error Map & Integrated & NonReg & $u^2$-MDN  & HSIFN &  Ours
  \end{tabular}
 \end{minipage}
 \hspace{22mm}
 \begin{minipage}[l]{0.1\linewidth}
 \begin{flushright}
    \vspace{-5mm}
     \includegraphics[width=0.54\linewidth]{figure_color_bar.png}
 \end{flushright}
 \end{minipage}

 \hspace{2mm}
 \begin{minipage}[l]{0.22\linewidth}
\begin{flushleft}
 \vspace{-0.1mm}
  \hspace{-3mm}
\includegraphics[width=1.045\linewidth]{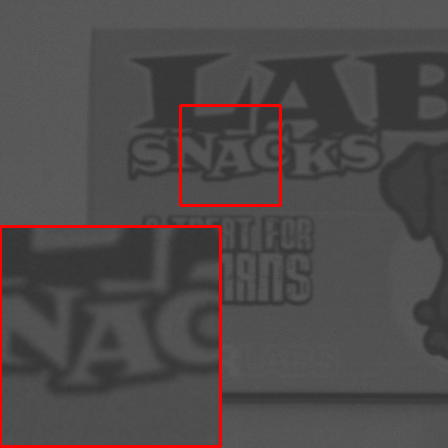}\\
   \begin{tabular}[c]{@{}c@{}}\hspace{7mm}Ground Truth\end{tabular}
  \end{flushleft}
 \end{minipage}
 \hspace{-1mm}
 \begin{minipage}[t]{0.5\linewidth}
  \begin{tabular}{cccccc}
   \includegraphics[width=0.2\linewidth]{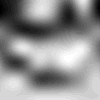}
  & \includegraphics[width=0.2\linewidth]{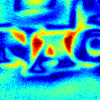}
  & \includegraphics[width=0.2\linewidth]{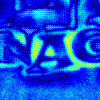}
  & \includegraphics[width=0.2\linewidth]{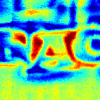}
  & \includegraphics[width=0.2\linewidth]{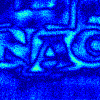}
  & \includegraphics[width=0.2\linewidth]{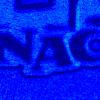}
   \\
    Input ($\times 16$) & BiQRNN & SSPSR  & MCNet & ESSA  & Optimized \\
   
 \includegraphics[width=0.2\linewidth]{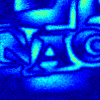}
  &\includegraphics[width=0.2\linewidth]{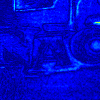}
  &\includegraphics[width=0.2\linewidth]{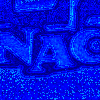}
  &\includegraphics[width=0.2\linewidth]{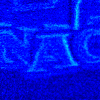}
  &\includegraphics[width=0.2\linewidth]{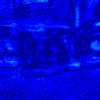}
  &\includegraphics[width=0.2\linewidth]{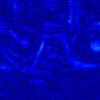}
  
  \\
   Error Map & Integrated & NonReg & $u^2$-MDN  & HSIFN &  Ours
  \end{tabular}
 \end{minipage}
 \hspace{22mm}
 \begin{minipage}[l]{0.1\linewidth}
 \begin{flushright}
    \vspace{-5mm}
     \includegraphics[width=0.54\linewidth]{figure_color_bar.png}
 \end{flushright}
 \end{minipage}

 \caption{Visual comparison on the Real dataset. The ground truth and input HSIs are shown with band 20. The results of different methods under scale factor $\times 4$, $\times 8$, $\times 16$ from top to bottom with the error maps. Zoom in for details}
 \label{fig:real visual results} 
\end{figure*}

\begin{figure}
\centering
\includegraphics[width=1\linewidth]{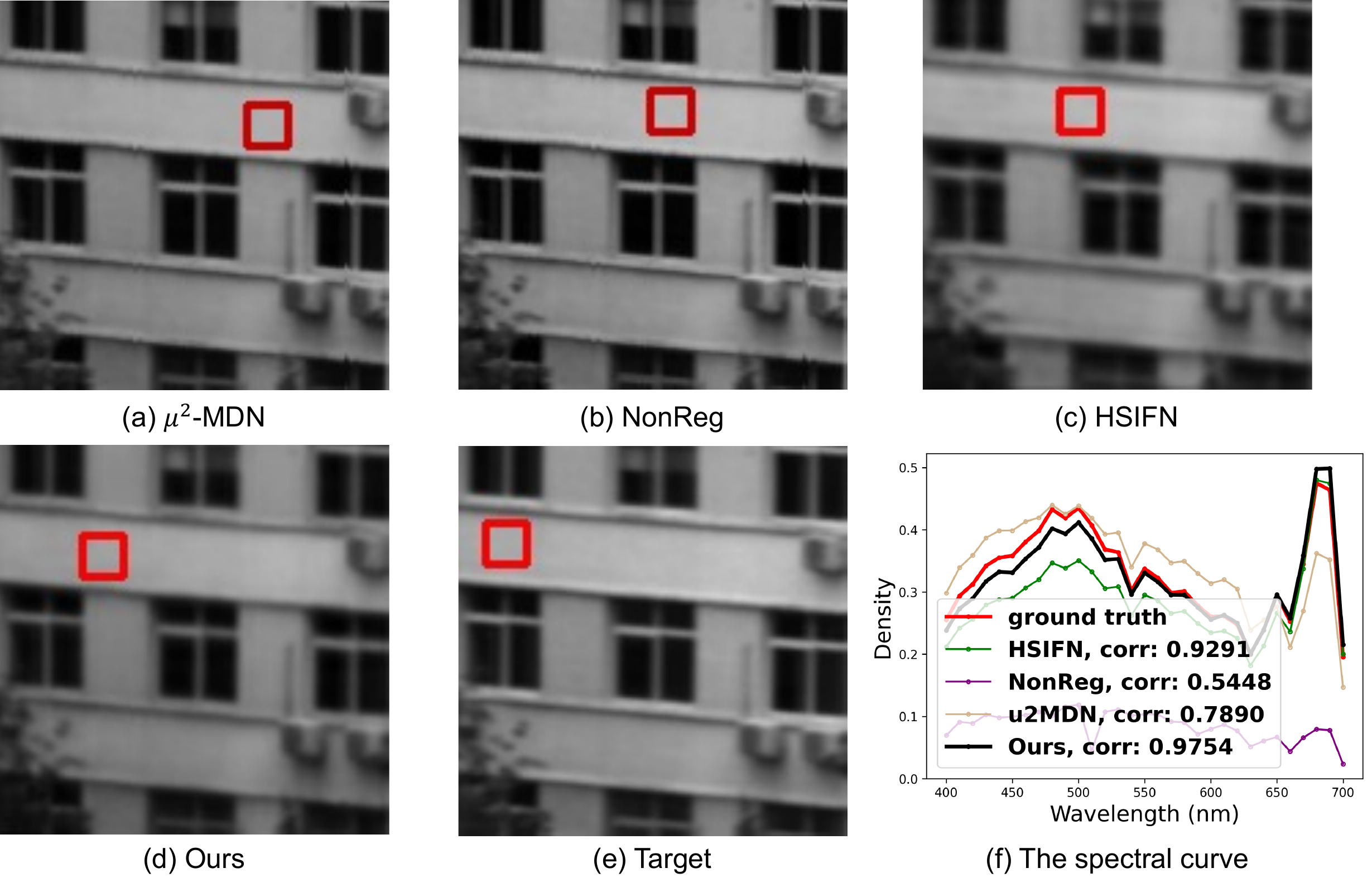}
   \caption{Comparison of spectral curves for the same material at different positions. (a)-(d) show the SR HSI generated by different methods. (f) presents the comparison of spectral curves, statistically derived from the selected pixels in the red box of each HSI}
\label{fig:spectral_comparison}
\end{figure}

\begin{table*}[htb]
\begin{center}
\small
\setlength{\tabcolsep}{0.11cm}
\renewcommand\arraystretch{0.5}
\caption{Quantitative comparison of various methods under several scale factors on the remote-sensing dataset, Chikusei. PSNR is in dB}
\label{tab:result on chikusei dataset}
\begin{tabular}{ccccccccccccc}
\toprule
\rowcolor{graycolor}

\makecell{Scale \\ Factor} & \makecell{Metric}                        & \makecell{Bicubic}    & \makecell{BiQRNN \\ (JSTAR'21)}  & \makecell{SSPSR \\ (TCI'20)} & \makecell{MCNet \\ (RS'20)}  & \makecell{ESSA \\ (ICCV'23)}   & \makecell{Optimized \\ (CVPR'19)} & \makecell{Integrated \\ (TGRS'19)}  & \makecell{NonReg \\ (TGRS'22)} & \makecell{$u^2$-MDN \\ (TGRS'22)} & \makecell{HSIFN \\ (TNNLS'24)} &  \makecell{Ours}      \\ \midrule
\multirow{3}{*}{4}
    & PSNR$\uparrow$ & 36.96 & 38.02 & 40.22 & 39.63 & 39.02 & 37.88 & 37.87 & 35.75 & 36.77 & 41.67 & \textbf{44.24} \\ & \\
    & SSIM$\uparrow$ & 0.945 & 0.954 & 0.972 & 0.968 & 0.966 & 0.959 & 0.957 & 0.958  & 0.949 & 0.984 & \textbf{0.989} \\ & \\
    & SAM$\downarrow$  & 0.060 & 0.066 & 0.049 & 0.050 & 0.064 & 0.253 & 0.056 & 0.213  & 0.078 & 0.056 & \textbf{0.044} \\ \midrule
\multirow{3}{*}{8}    
	& PSNR$\uparrow$ & 33.99 & 35.43 & 36.06 & 35.79 & 35.52 & 38.35 & 35.41 & 35.63 & 36.72 & 39.48 & \textbf{41.99} \\ & \\
    & SSIM$\uparrow$ & 0.910 & 0.927 & 0.935 & 0.932 & 0.930 & 0.962 & 0.937 & 0.957  & 0.948 & 0.974 & \textbf{0.984} \\ & \\
    & SAM$\downarrow$  & 0.086 & 0.076 & 0.069 & 0.076 & 0.086 & 0.234 & 0.076 & 0.216  & 0.079 & 0.066 & \textbf{0.050} \\ \midrule
\multirow{3}{*}{16}    
	& PSNR$\uparrow$ & 31.64 & 32.60 & 33.31 & 33.19 & 32.83 & 36.82 & 32.50 & 34.60 & 36.78 & 31.85 & \textbf{38.90} \\ & \\
    & SSIM$\uparrow$ & 0.879 & 0.886 & 0.902 & 0.900 & 0.894 & 0.952 & 0.904 & 0.949  & 0.944 & 0.885 & \textbf{0.968} \\ & \\
    & SAM$\downarrow$  & 0.117 & 0.113 & 0.100 & 0.102 & 0.118 & 0.273 & 0.104 & 0.244  & 0.082 & 0.107 & \textbf{0.065} \\

\bottomrule
\end{tabular}
\end{center}
\end{table*}

\begin{figure*}
 \centering
\small
\hspace{1mm}
 \setlength{\tabcolsep}{0.1cm}
\begin{minipage}[l]{0.22\linewidth}
\begin{flushleft}
 \vspace{-0.1mm}
  \hspace{-3mm}
\includegraphics[width=1.045\linewidth]{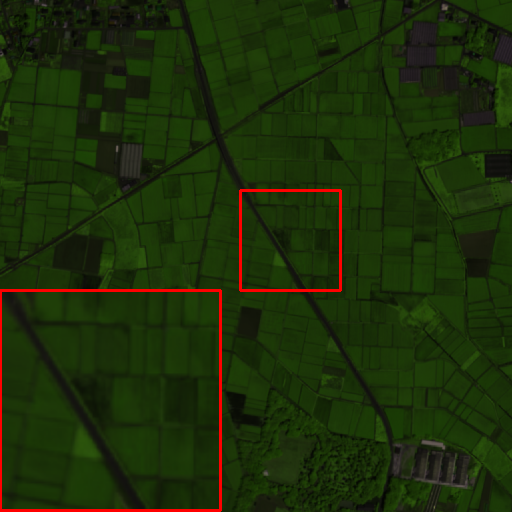}\\
   \begin{tabular}[c]{@{}c@{}}\hspace{7mm}Ground Truth\end{tabular}
  \end{flushleft}
 \end{minipage}
 \hspace{-1mm}
 \begin{minipage}[t]{0.5\linewidth}
  \begin{tabular}{cccccc}
   \includegraphics[width=0.2\linewidth]{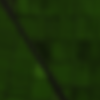}
  & \includegraphics[width=0.2\linewidth]{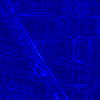}
  & \includegraphics[width=0.2\linewidth]{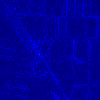}
  & \includegraphics[width=0.2\linewidth]{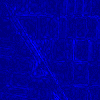}
  & \includegraphics[width=0.2\linewidth]{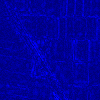}
  & \includegraphics[width=0.2\linewidth]{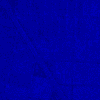}
   \\
    Input ($\times 4$) & BiQRNN & SSPSR  & MCNet & ESSA  & Optimized \\
   
 \includegraphics[width=0.2\linewidth]{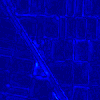}
  &\includegraphics[width=0.2\linewidth]{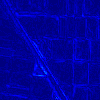}
  &\includegraphics[width=0.2\linewidth]{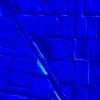}
  &\includegraphics[width=0.2\linewidth]{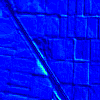}
  &\includegraphics[width=0.2\linewidth]{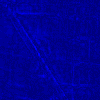}
  &\includegraphics[width=0.2\linewidth]{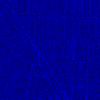}
  
  \\      
   Error Map & Integrated & NonReg & $u^2$-MDN  & HSIFN &  Ours
  \end{tabular}
 \end{minipage}
 \hspace{22mm}
 \begin{minipage}[l]{0.1\linewidth}
 \begin{flushright}
    \vspace{-5mm}
     \includegraphics[width=0.54\linewidth]{figure_color_bar.png}
 \end{flushright}
 \end{minipage}

\hspace{2mm}
 \begin{minipage}[l]{0.22\linewidth}
\begin{flushleft}
 \vspace{-0.1mm}
  \hspace{-3mm}
\includegraphics[width=1.045\linewidth]{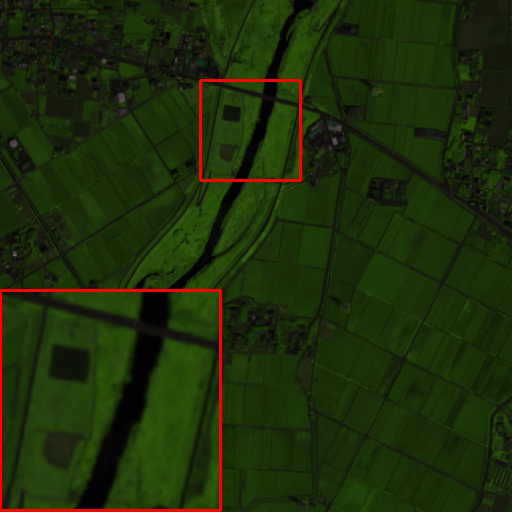}\\
   \begin{tabular}[c]{@{}c@{}}\hspace{7mm}Ground Truth\end{tabular}
  \end{flushleft}
 \end{minipage}
 \hspace{-1mm}
 \begin{minipage}[t]{0.5\linewidth}
  \begin{tabular}{cccccc}
   \includegraphics[width=0.2\linewidth]{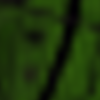}
  & \includegraphics[width=0.2\linewidth]{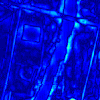}
  & \includegraphics[width=0.2\linewidth]{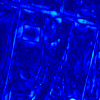}
  & \includegraphics[width=0.2\linewidth]{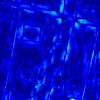}
  & \includegraphics[width=0.2\linewidth]{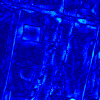}
  & \includegraphics[width=0.2\linewidth]{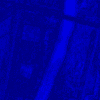}
   \\
    Input ($\times 8$) & BiQRNN & SSPSR  & MCNet & ESSA  & Optimized \\
   
 \includegraphics[width=0.2\linewidth]{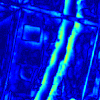}
  &\includegraphics[width=0.2\linewidth]{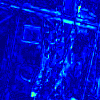}
  &\includegraphics[width=0.2\linewidth]{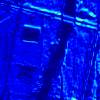}
  &\includegraphics[width=0.2\linewidth]{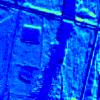}
  &\includegraphics[width=0.2\linewidth]{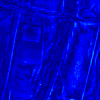}
  &\includegraphics[width=0.2\linewidth]{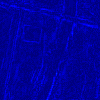}
  
  \\
   Error Map & Integrated & NonReg & $u^2$-MDN  & HSIFN &  Ours
  \end{tabular}
 \end{minipage}
 \hspace{22mm}
 \begin{minipage}[l]{0.1\linewidth}
 \begin{flushright}
    \vspace{-5mm}
     \includegraphics[width=0.54\linewidth]{figure_color_bar.png}
 \end{flushright}
 \end{minipage}

 \hspace{2mm}
 \begin{minipage}[l]{0.22\linewidth}
\begin{flushleft}
 \vspace{-0.1mm}
  \hspace{-3mm}
\includegraphics[width=1.045\linewidth]{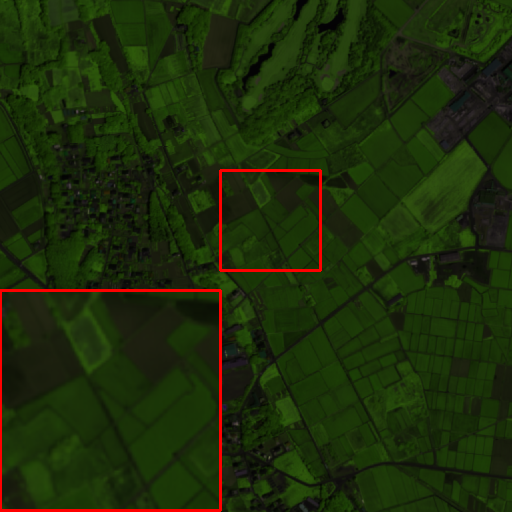}\\
   \begin{tabular}[c]{@{}c@{}}\hspace{7mm}Ground Truth\end{tabular}
  \end{flushleft}
 \end{minipage}
 \hspace{-1mm}
 \begin{minipage}[t]{0.5\linewidth}
  \begin{tabular}{cccccc}
   \includegraphics[width=0.2\linewidth]{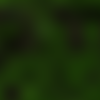}
  & \includegraphics[width=0.2\linewidth]{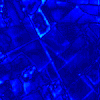}
  & \includegraphics[width=0.2\linewidth]{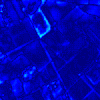}
  & \includegraphics[width=0.2\linewidth]{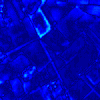}
  & \includegraphics[width=0.2\linewidth]{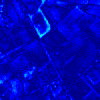}
  & \includegraphics[width=0.2\linewidth]{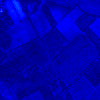}
   \\
    Input ($\times 16$) & BiQRNN & SSPSR  & MCNet & ESSA  & Optimized \\
   
 \includegraphics[width=0.2\linewidth]{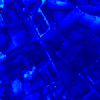}
  &\includegraphics[width=0.2\linewidth]{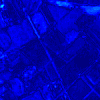}
  &\includegraphics[width=0.2\linewidth]{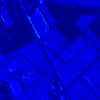}
  &\includegraphics[width=0.2\linewidth]{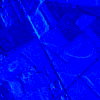}
  &\includegraphics[width=0.2\linewidth]{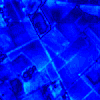}
  &\includegraphics[width=0.2\linewidth]{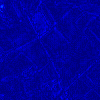}
  
  \\
   Error Map & Integrated & NonReg & $u^2$-MDN  & HSIFN &  Ours
  \end{tabular}
 \end{minipage}
 \hspace{22mm}
 \begin{minipage}[l]{0.1\linewidth}
 \begin{flushright}
    \vspace{-5mm}
     \includegraphics[width=0.54\linewidth]{figure_color_bar.png}
 \end{flushright}
 \end{minipage}

 \caption{Visual comparison on the Chikusei dataset. Bands 70-100-36 are visualized as the R-G-B channels. The results of different methods under scale factor $\times 4$, $\times 8$, $\times 16$ from top to bottom with the error maps. Zoom in for details}
 \label{fig:chikusei visual results} 
\end{figure*}

\noindent{\textbf{Results on the Real Dataset.}}
In Table~\ref{tab:result on real dataset} and Fig.~\ref{fig:real visual results}, we provide quantitative and qualitative comparisons with other state-of-the-art methods.
Different from the Flower dataset, the Real dataset has less paired data, but a larger misalignment. Thus, we can observe that most methods have a performance gain in three metrics over Bicubic, but this gain is very limited when compared to the Flower dataset.
Compared with other reference-based methods, our method can effectively perform alignment, aggregation, and fusion at high-scale factors $\times8$ and $\times16$. Especially, our method has a 4 dB gain in PSNR over the suboptimal method, HSIFN \citep{lai2024hyperspectral} at factor of $\times8$ shown in Table~\ref{tab:result on real dataset}.
The visual results in Fig.~\ref{fig:real visual results} also demonstrate the ability of our model to generate high-quality HR HSI with finer details, \ie, fewer errors with ground truth. The blurred results of SISR methods indicate that they are not able to generate detailed information, \ie, high-frequency information, efficiently.
Additionally, to mitigate the influence of spatial misalignment in the comparison, which could result in lower metrics relative to the reference, we select different but similar regions from the SR results and the target HSI for spectral curve comparison, as shown in Fig.~\ref{fig:spectral_comparison}. In the same HSI image, the spectral curves from different similar regions should exhibit high similarity. It can be seen that our method demonstrates the highest spectral similarity.
For example, we can see the result of the "book" scene at the factor of $\times8$, which is unaligned with the ground truth, \ie, higher intensity in the error map, resulting in bad performance.

\noindent{\textbf{Results on the Chikusei Dataset.}}
In addition to the natural HSI dataset, we adopt a remote-sensing dataset, \ie, Chikusei to evaluate our method with others. 
As shown in Table~\ref{tab:result on chikusei dataset} and Fig.~\ref{fig:chikusei visual results}, our method outperforms others on quantitative metrics and visualization results.
Most of the methods perform similarly to those on the other two datasets, and the inaccurate alignment rather affects the performance of the fusion-based methods resulting in low-quality evaluation metrics and visualization results.
To maintain spectral information of input HSI and build a trainable model on a single GPU, we adopt a two-level architecture of our proposed method, \ie, scale level $L=2$, and set the output channel of the first layer of HSI encoder to $C=128$ on the Chikusei dataset. Yet, in the Flower and Real dataset, we set it to $C=64$. From Table~\ref{tab:result on chikusei dataset}, our method achieves at least 2 dB in PSNR over other suboptimal methods. Especially, the performance of HSIFN at $\times16$ drops off greatly, which is mainly due to the poor results of the alignment, resulting in the network's inability to learn useful information from the reference image. It seriously affects the final fusion, which is only a little better than Bicubic.
To further evaluate and measure our model, we provide the comparison of visual results in Fig.~\ref{fig:chikusei visual results}. 
We magnify the area within the red box across three HSIs under varying scale factors and present their error maps compared to the ground truth.
It can be seen that our method has the smallest intensity on the overall error map, which means that our method closely approximates the ground truth and has higher quality compared to other methods.

\subsection{Ablation Study}
\label{sec: ablation}

In this section, we conduct ablation studies to verify the effectiveness of each module in our approach and a specific analysis of each module with metrics or visualization plots, including the Two-Stage Image Alignment (TSIA) module, Feature Aggregation (FA) module, and Attention Fusion (AF) module.

\begin{table}[]
   \setlength{\tabcolsep}{0.31cm}
       \centering
       \caption{Break-down ablation studies to verify the effectiveness of modules about \textbf{T}wo-\textbf{S}tage \textbf{I}mage \textbf{A}lignment (\textbf{TSIA}) module, \textbf{F}eature \textbf{A}ggregation (\textbf{FA}) module, \textbf{A}ttention \textbf{F}usion (\textbf{AF}) module, \textbf{O}ptical-\textbf{F}low (\textbf{OF}) model, and \textbf{W}arp\textbf{N}et (\textbf{WN}). Results on the Real dataset at scale factor $\times8$}
       \label{tab:ablation-modules}
        \begin{tabular}{ccccccc}
        \hline
        \rowcolor{graycolor}
        \multicolumn{2}{c}{\textbf{TSIA}} & & & & & \\
        	\rowcolor{graycolor}
        \textbf{OF} & \textbf{WN}  & \multirow{-2}{*}{\textbf{FA}} & \multirow{-2}{*}{\textbf{AF}} & \multirow{-2}{*}{PSNR$\uparrow$} & \multirow{-2}{*}{SSIM$\uparrow$} & \multirow{-2}{*}{SAM$\downarrow$} \\
        \hline
        	- & - & - & - & 31.48 & 0.929 & 0.101 \\
        \checkmark & \checkmark & - & - & 34.74 & 0.954 & 0.068 \\
        \checkmark & \checkmark & \checkmark & - & 35.81 & 0.963 & 0.047 \\
        - & - & \checkmark & \checkmark & 36.66 & 0.967 & 0.045 \\
        	\checkmark & - & \checkmark & \checkmark & 37.12 & 0.973 & 0.044\\
        \checkmark & \checkmark & \checkmark & \checkmark & \textbf{37.37} & \textbf{0.974} & \textbf{0.042} \\
        \hline
        \end{tabular}
\end{table}

\begin{table}[]
   \setlength{\tabcolsep}{0.21cm}
       \centering
       \caption{Ablation study on the two-stage image alignment (TSIA) module. Results on the optical flow dataset, and Real dataset at scale factor $\times8$}
       \label{tab:ablation-TSIA}
        \begin{tabular}{cc|cccc}
        \hline
        \rowcolor{graycolor}
         \textbf{Optical flow} & \textbf{WarpNet} & EPE$\downarrow$ & F1$\downarrow$ & PSNR$\uparrow$ & SSIM$\uparrow$ \\ \hline
        Pre-trained & - & 11.86 & 41.52 & 26.17 & 0.827 \\
        Fine-tuned & \checkmark & \textbf{1.77} & \textbf{16.84} & \textbf{34.46} & \textbf{0.955} \\
        \hline
        \end{tabular}
\end{table}

\begin{table}[]
   \setlength{\tabcolsep}{0.36cm}
       \centering
       \caption{Ablation study on the feature aggregation (FA) module. Results on the Real dataset at scale factor $\times8$}
       \label{tab:ablation-FA}
        \begin{tabular}{c|cccc}
        \hline
        \rowcolor{graycolor}
        \textbf{Methods} & Layers & PSNR$\uparrow$ & SSIM$\uparrow$ & SAM$\downarrow$ \\ \hline
        w/o interaction & 2 & 35.64 & 0.960 & 0.047 \\
        w/ interaction & 2 & \textbf{37.19} & \textbf{0.973} & \textbf{0.042} \\ \hline
        w/o interaction & 3 &  37.25 & 0.974 & 0.043 \\
        w/ interaction & 3 & \textbf{37.37} & \textbf{0.974} & \textbf{0.042} \\
        \hline
        \end{tabular}
\end{table}

\subsubsection{Effect of Two-Stage Image Alignment Module}

Alignment is essential for further aggregation and fusion, so we propose a two-stage image alignment (TSIA) module, consisting of the optical flow model and the WarpNet model shown in Fig.~\ref{fig:optical-flow}. 
As shown in Tables~\ref{tab:ablation-modules} and~\ref{tab:ablation-TSIA}, we evaluate the comprehensive TSIA module along with its two sub-models, the optical flow model and WarpNet. The baseline is the UNet \citep{ronneberger2015u} listed in the first row (UNet-only). The UNet configured with concatenated LR HSI and HR RGB inputs achieves a PSNR of 31.48 dB and an SSIM of 0.929.
To assess the full TSIA module, we incorporate TSIA into the UNet-only architecture as shown in the second row. Within TSIA, the base model gains by 3.26 dB and 0.025 in PSNR and SSIM, respectively, which demonstrates that TSIA effectively aligns the HR RGB reference and enhances the fusion process to improve performance. Furthermore, we analyze the impact of TSIA by combining it with other modules (FA and AF). 
It can be seen that adding OF before FA and AF results in a 0.46 dB improvement over using FA and AF alone, which achieves a PSNR of 36.66 dB. Moreover, incorporating all components of TSIA, including OF and WN, yields the highest performance, with a PSNR of 37.37 dB, an SSIM of 0.974, and a SAM of 2.00. These results confirm that explicitly precise alignment with TSIA significantly enhances super-resolution results, facilitating the subsequent FA and AF modules to more effectively aggregate and integrate features for final high-resolution HSI super-resolution.

As shown in Table~\ref{tab:ablation-TSIA}, we conduct a quantitative evaluation of the TSIA, consisting of an optical flow model and WarpNet. The term \textit{Pre-trained} refers to the optical flow model, RAFT, which is trained on the FlyingChairs and FlyingThings datasets. The \textit{Fine-tuned} indicates that RAFT has been further trained using our pipeline with the generated cross-resolution dataset. Following the study \citep{teed2020raft}, we assess the performance of optical flow prediction using two metrics: average End-Point-Error (EPE) and an error rate (F1) where both absolute and relative errors exceed 3\% and 5\% respectively, across all pixels. The fine-tuned RAFT demonstrates an improvement in EPE from 11.86 to 1.77 and in F1 from 41.52 to 16.84, demonstrating that our pipeline enhances the effectiveness of cross-resolution prediction. To assess the warp component of TSIA, we compare traditional warp operation with WarpNet on the Real dataset, analyzing the alignment between warped HR reference RGB and target HR RGB images using the quality metrics PSNR and SSIM. The comparisons reveal marked improvements in both PSNR and SSIM with the use of WarpNet, confirming its efficacy in refining corrupted textures to more closely match the target and in improving the accuracy of explicit alignment.

As illustrated in Fig.~\ref{fig:RAFT-warp-warpNet}, we provide the warped results of one scene from the Real dataset. The relative position of the red box remains consistent across each image. Compared with Fig.~\ref{fig:RAFT-warp-warpNet}(a) and Fig.~\ref{fig:RAFT-warp-warpNet}(d), we can observe that there is an obvious misalignment and occlusion between target and reference images. 
The \textit{ft} in the figure stands for \textit{fine-tuned}, indicating that the optical flow model, RAFT, has been fine-tuned. Therefore, in Fig.~\ref{fig:RAFT-warp-warpNet}(b) and Fig.~\ref{fig:RAFT-warp-warpNet}(c), RAFT is not fine-tuned by our proposed pipeline but is merely pre-trained on traditional optical flow datasets, where the corrupted texture is evident.
In contrast, Fig.~\ref{fig:RAFT-warp-warpNet}(e) and Fig.~\ref{fig:RAFT-warp-warpNet}(f) depict images where RAFT is fine-tuned and WarpNet is applied subsequent to the optical flow prediction. This combination leads to significant improvements, particularly evident in the attenuation of corrupted textures. The additional application of WarpNet further refines the textures to more closely resemble the target.
Consequently, both the quantitative and visual results demonstrate that our proposed Two-Stage Image Alignment (TSIA) can more accurately align images and effectively refine image details.

\begin{figure}
\begin{center}
\setlength{\tabcolsep}{0.01cm}
    \begin{subfigure}[h]{0.15\textwidth}
          \centering
          \small
          \includegraphics[width=1\linewidth]{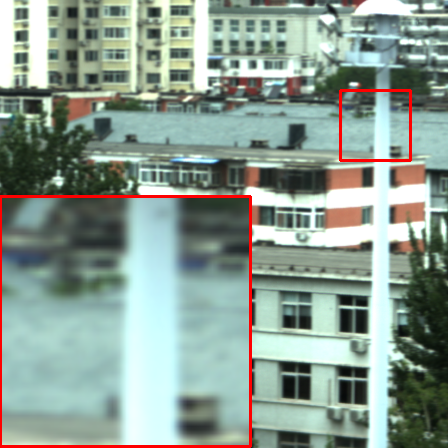}
          \caption{Target}
      \end{subfigure}
    \hfill
    \begin{subfigure}[h]{0.15\textwidth}
          \centering
          \small
          \includegraphics[width=1\linewidth]{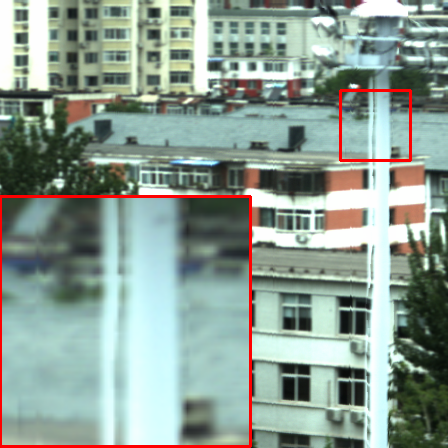}
          \caption{RAFT (w/o \textit{ft})}
      \end{subfigure}
    \hfill
    \begin{subfigure}[h]{0.15\textwidth}
          \centering
          \small
          \includegraphics[width=1\linewidth]{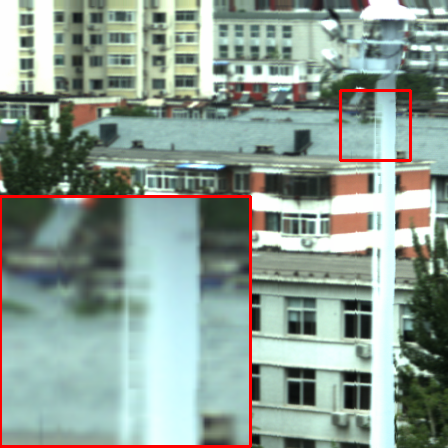}
          \caption{WarpNet (w/o \textit{ft})}
      \end{subfigure}
    \vfill
    \begin{subfigure}[h]{0.15\textwidth}
          \centering
          \small
          \includegraphics[width=1\linewidth]{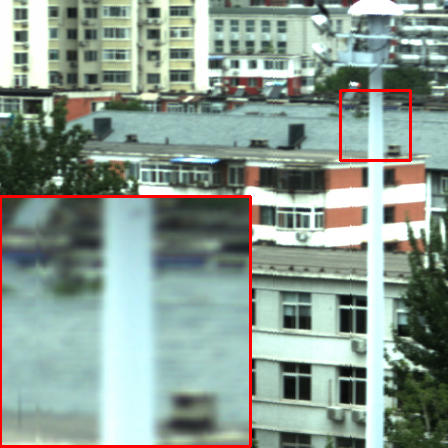}
          \caption{Reference}
      \end{subfigure}
    \hfill
    \begin{subfigure}[h]{0.15\textwidth}
          \centering
          \small
          \includegraphics[width=1\linewidth]{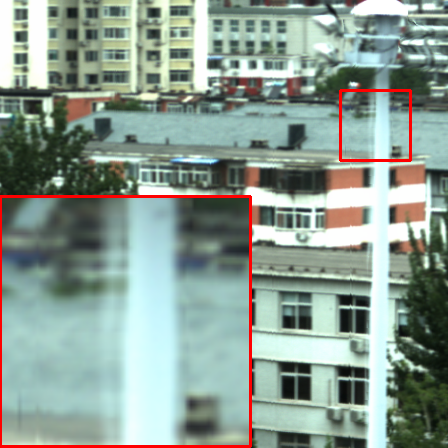}
          \caption{RAFT (w/ \textit{ft})}
      \end{subfigure}
    \hfill
    \begin{subfigure}[h]{0.15\textwidth}
          \centering
          \small
          \includegraphics[width=1\linewidth]{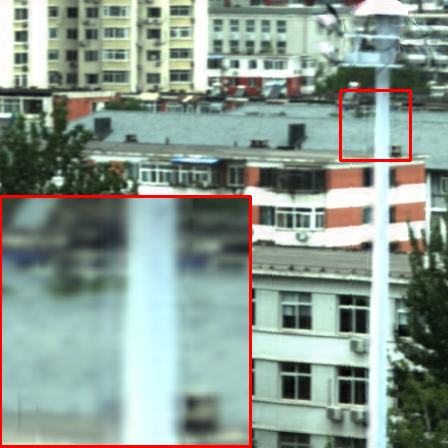}
          \caption{WarpNet (w/ \textit{ft})}
      \end{subfigure}
\end{center}
   \caption{The warped HR RGB images at scale factor $\times4$ by RAFT and WarpNet. The term \textit{ft} denotes \textit{fine-tuned}, indicating that the optical flow model, RAFT, has been fine-tuned. (a) The target image. (b) The warped image by RAFT, which is pre-trained on traditional optical flow datasets (\eg, FlyingChairs and FlyingThings) \citep{teed2020raft}. (c) The warped image by WarpNet, but RAFT is held fixed as in (b). (d) The reference image. (e) The warped image by RAFT, which is fine-tuned through our proposed pipeline. (f) The warped image by WarpNet, where RAFT is fine-tuned as in (e)}
    \label{fig:RAFT-warp-warpNet}
\end{figure}

\begin{figure}
\begin{center}
\setlength{\tabcolsep}{0.01cm}
    \begin{subfigure}[h]{0.15\textwidth}
          \centering
          \small
          \includegraphics[width=1\linewidth]{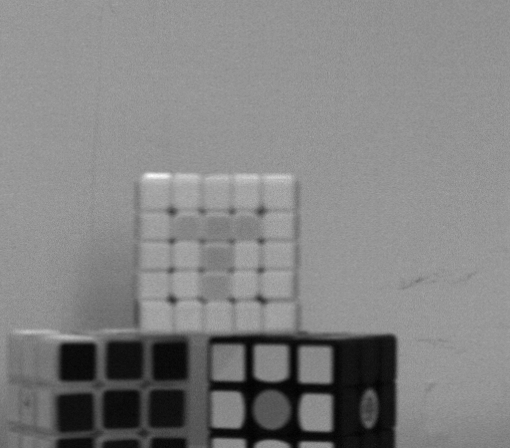}
          \caption{Target}
      \end{subfigure}
    \hfill
    \begin{subfigure}[h]{0.15\textwidth}
          \centering
          \small
          \includegraphics[width=1\linewidth]{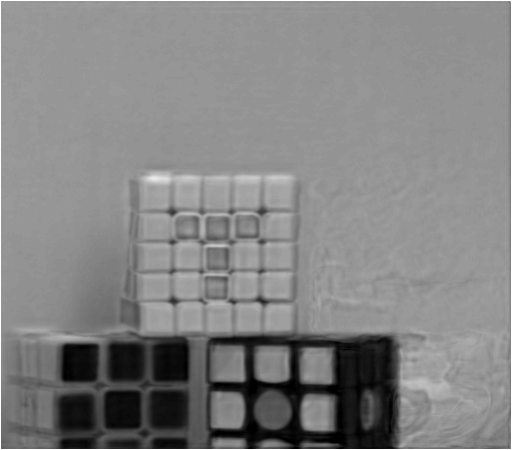}
          \caption{HSIFN}
      \end{subfigure}
    \hfill
    \begin{subfigure}[h]{0.15\textwidth}
          \centering
          \small
          \includegraphics[width=1\linewidth]{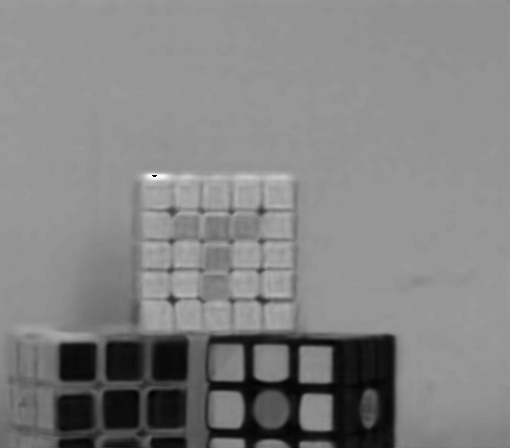}
          \caption{Ours}
      \end{subfigure}
    \vfill
    \begin{subfigure}[h]{0.15\textwidth}
          \centering
          \small
          \includegraphics[width=1\linewidth]{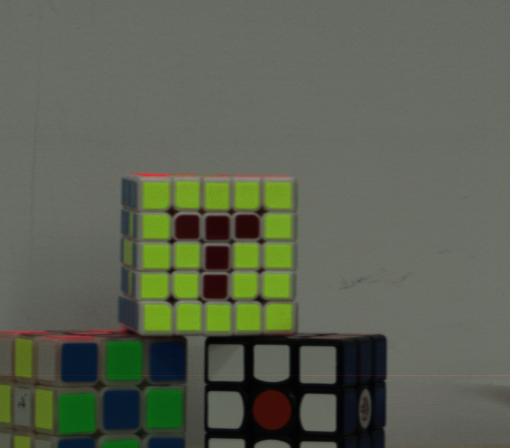}
          \caption{Reference ($\times 8$)}
      \end{subfigure}
    \hfill
    \begin{subfigure}[h]{0.15\textwidth}
          \centering
          \small
          \includegraphics[width=1\linewidth]{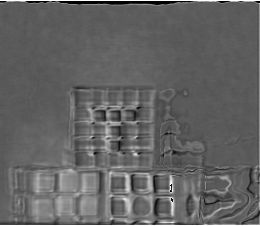}
          \caption{Feature Map}
      \end{subfigure}
    \hfill
    \begin{subfigure}[h]{0.15\textwidth}
          \centering
          \small
          \includegraphics[width=1\linewidth]{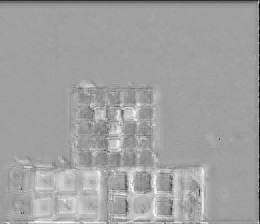}
          \caption{Feature Map}
      \end{subfigure}
\end{center}
   \caption{The comparison of feature maps. HSIFN \citep{lai2024hyperspectral} is based on optical flow. (a) The target image shows the 20th band of the ground-truth HSI. (b) The result of HSIFN. (c) The result of our method is at level $l==2$. (d) The reference image. We can see the spatial misalignment between target and reference images. (e) The feature map of HSIFN. (f) The feature map of our method. Please zoom in for better visualization}
    \label{fig:Ablation DCN}
\end{figure}

\subsubsection{Effect of Feature Aggregation Module}

Feature aggregation (FA) can be regarded as alignment at the feature level but unlike the previous alignment at the image level. It not only adjusts the matching of features at the corresponding positions but also further aggregates globally similar features, to construct more effective reference features, which can assist in the subsequent fusion of high-quality HR HSI. As shown in Table~\ref{tab:ablation-modules},  the network with TSIA and FA gains 1.07 dB and 0.09 in PSNR and SSIM over single with TSIA. Besides, there is a large improvement in the SAM metrics (0.02 drop), indicating that FA can aggregate effective features to reconstruct the original spectral relationships.
To verify the effectiveness of the interaction between the FA module with further attention fusion module, we conduct an ablation study in Table~\ref{tab:ablation-FA}. 
Since our AF module performs dynamic aggregation based on DCN, it needs to learn an offset to guide the module in aggregating the information at a specific location, so interaction here specifically means that the AF module utilizes the results of the fusion module to guide the learning of the offset, and iteratively adjusts according to the fusion results. As shown in Table~\ref{tab:ablation-FA}, we evaluate two base networks with different scales. It can be seen that 'w/ interaction' with different scales can contribute to performance improvement. Specifically, the gain on all metrics is greater at $Layers=2$, indicating that the effectiveness of feature aggregation can be better improved by interaction with the fusion module when the number of parameters is small, which verifies the effectiveness of our method.

As shown in Fig.~\ref{fig:Ablation DCN}, we provide the results and feature maps of HSIFN and ours. HSIFN is based on optical flow to align for multi-scale features and then adopt an attention module to fine-tune features. However, this explicit alignment model (\eg, optical flow model) without specific training suffers in feature effectiveness, and even attention modules based on residual convolution struggle to get clean feature maps well. Our method based on IDFA in the FA module can better aggregate cleaner features under interaction (\ie, the fusion result guidance) with the fusion module.

\begin{table}[]
   \setlength{\tabcolsep}{0.54cm}
       \centering
       \caption{Ablation study on the attention fusion (AF) module. Results on the Real dataset at scale factor $\times8$}
       \label{tab:ablation-AF}
        \begin{tabular}{c|cccc}
        \hline
        \rowcolor{graycolor}
        \textbf{Methods} & PSNR$\uparrow$ & SSIM$\uparrow$ & SAM$\downarrow$ \\ \hline
        w/ UNet & 35.81 & 0.963 & 0.047 \\
        w/ SSA & 36.06 & 0.964 & 0.045 \\
        w/ SSA+RCA & \textbf{37.37} & \textbf{0.974} & \textbf{0.042} \\
        \hline
        \end{tabular}
\end{table}

\begin{table}[]
    \setlength{\tabcolsep}{0.2cm}
       \centering
		\caption{Comparison of different scale levels in our model. Results on the Real dataset at resolution of $256\times256$ and scale factor $\times8$ with an NVIDIA 3090 GPU. The comparisons are drawn in Fig.~\ref{fig:psnr_ssim_params_flops}}
		\label{tab:ablation-scales}
        \begin{tabular}{c|ccccc}
        \hline
        \rowcolor{graycolor}
		\textbf{Method} & Layers & Params/M & FLOPs/G & Time/ms & PSNR$\uparrow$\\ \hline
		BiQRNN & - & 1.30 & 468.7 & 156.6 & 31.66 \\
        	SSPSR & - & 20.82 & 180.4 & 78.1 & 31.13 \\
        	MCNet & - & 2.20 & 280.3 & 116.8 & 31.31\\
        	ESSA & - & 11.10 & 78.14 & 42.1 & 31.85 \\
        	Optimized & - & 0.20 & 14.17 & 16.4 & 26.99 \\
        	NonReg & - & 2.00 & 161.61 & - & 30.00 \\
        	$u^2$-MDN & - & 0.06 & - & - & 30.18 \\
        HSIFN & - & 21.01 & 594.1 & 167.8 & 33.13 \\
        Ours-S & 1 & 10.97 & 165.6 & 63.9 & 35.16 \\
        Ours-M & 2 & 13.90 & 219.1 & 85.8 & 37.19 \\
        Ours-L & 3 & 26.62 & 276.7 & 103.5 & \textbf{37.37} \\
        \hline
        \end{tabular}
\end{table}

\begin{figure}
\begin{center}
\setlength{\tabcolsep}{0.01cm}
    \begin{subfigure}[h]{0.15\textwidth}
          \centering
          \small
          \includegraphics[width=1\linewidth]{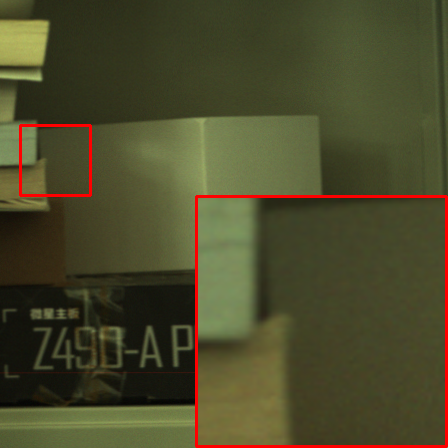}
          \caption{Target}
      \end{subfigure}
    \hfill
    \begin{subfigure}[h]{0.15\textwidth}
          \centering
          \small
          \includegraphics[width=1\linewidth]{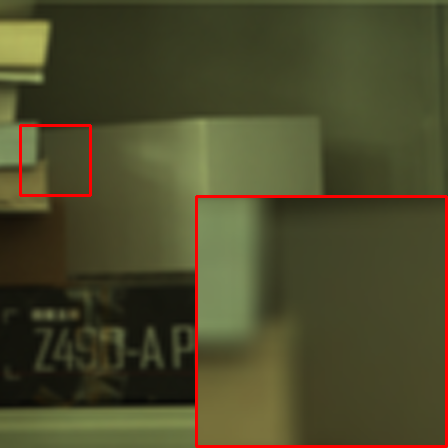}
          \caption{Scale Factor $\times4$}
      \end{subfigure}
    \hfill
    \begin{subfigure}[h]{0.15\textwidth}
          \centering
          \small
          \includegraphics[width=1\linewidth]{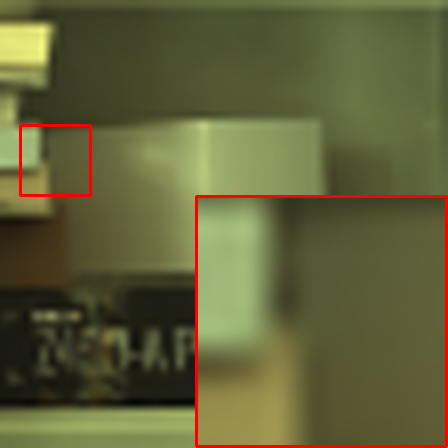}
          \caption{Scale Factor $\times8$}
      \end{subfigure}
    \vfill
    \begin{subfigure}[h]{0.15\textwidth}
          \centering
          \small
          \includegraphics[width=1\linewidth]{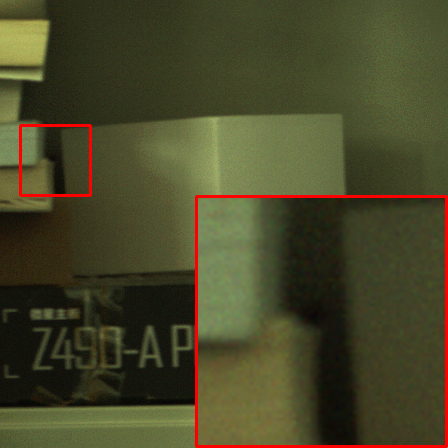}
          \caption{Reference}
      \end{subfigure}
    \hfill
    \begin{subfigure}[h]{0.15\textwidth}
          \centering
          \small
          \includegraphics[width=1\linewidth]{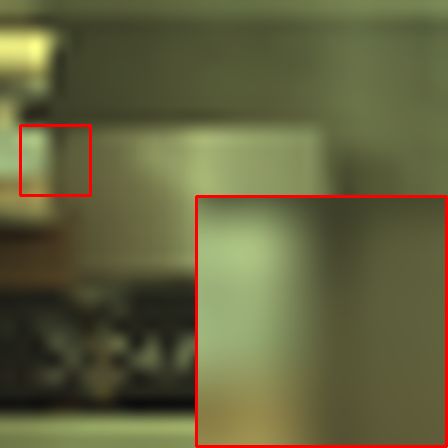}
          \caption{Scale Factor $\times16$}
      \end{subfigure}
    \hfill
    \begin{subfigure}[h]{0.15\textwidth}
          \centering
          \small
          \includegraphics[width=1\linewidth]{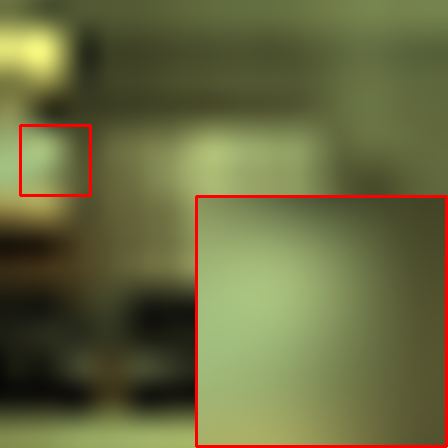}
          \caption{Scale Factor $\times32$}
      \end{subfigure}
\end{center}
   \caption{Comparison of different scale factors. We use Gaussian blur when downsampling at different scale factors, and then upsampling for presentation. Spatial detail can be seen at scale factors of $\times4$ and $\times8$. The result at $\times16$ blurs most of the detail and enlarges the center black area. The results at $\times32$ are completely indistinguishable from texture details}
    \label{fig:large scale}
\end{figure}

\begin{figure*}
\begin{center}
\setlength{\tabcolsep}{0.1cm}
    \begin{subfigure}[h]{0.33\textwidth}
          \centering
          \small
          \includegraphics[width=1\linewidth]{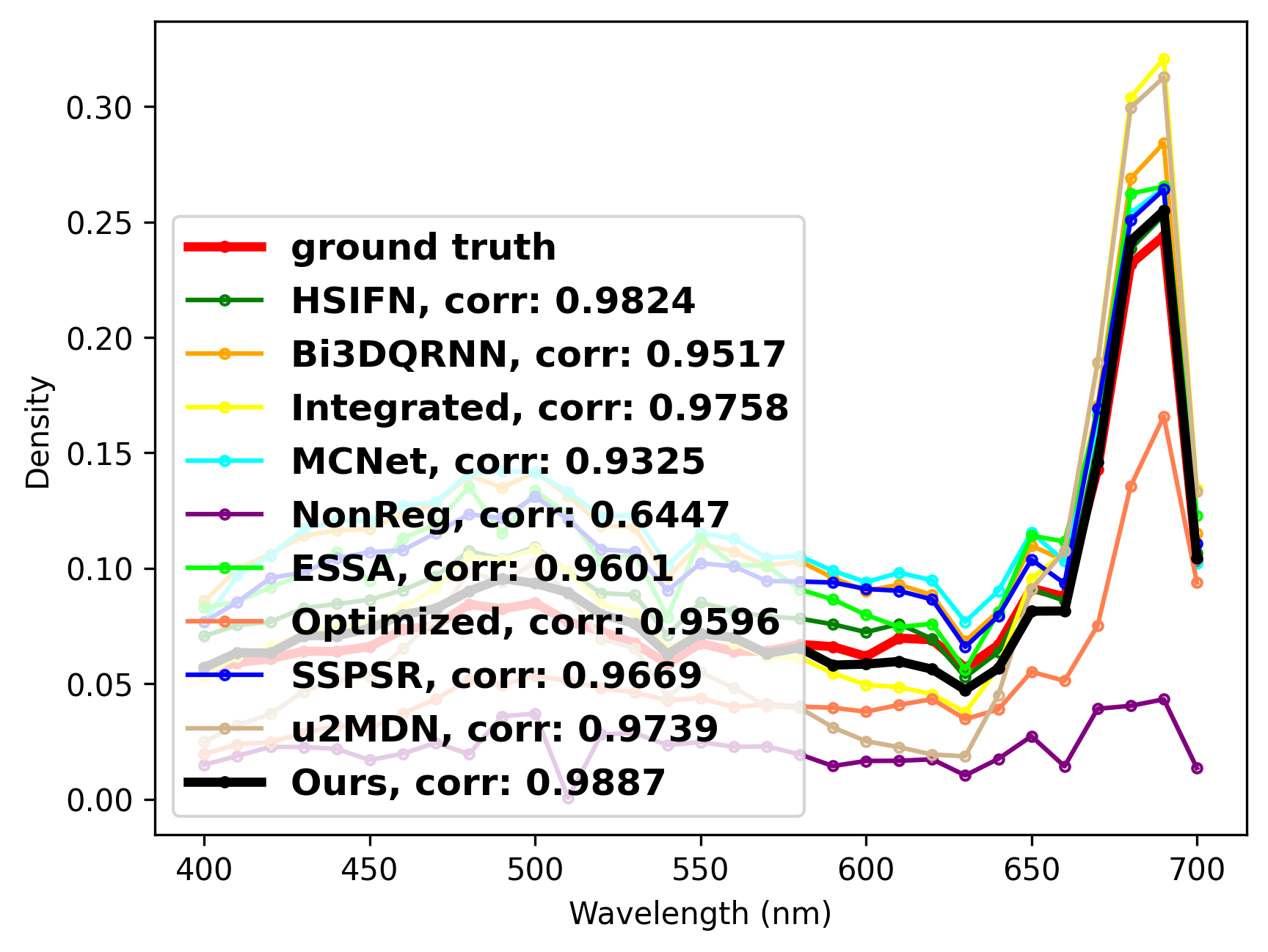}
          \caption{The spectral curve of some selected pixels.}
      \end{subfigure}
    \hfill
    \begin{subfigure}[h]{0.33\textwidth}
          \centering
          \small
          \includegraphics[width=1\linewidth]{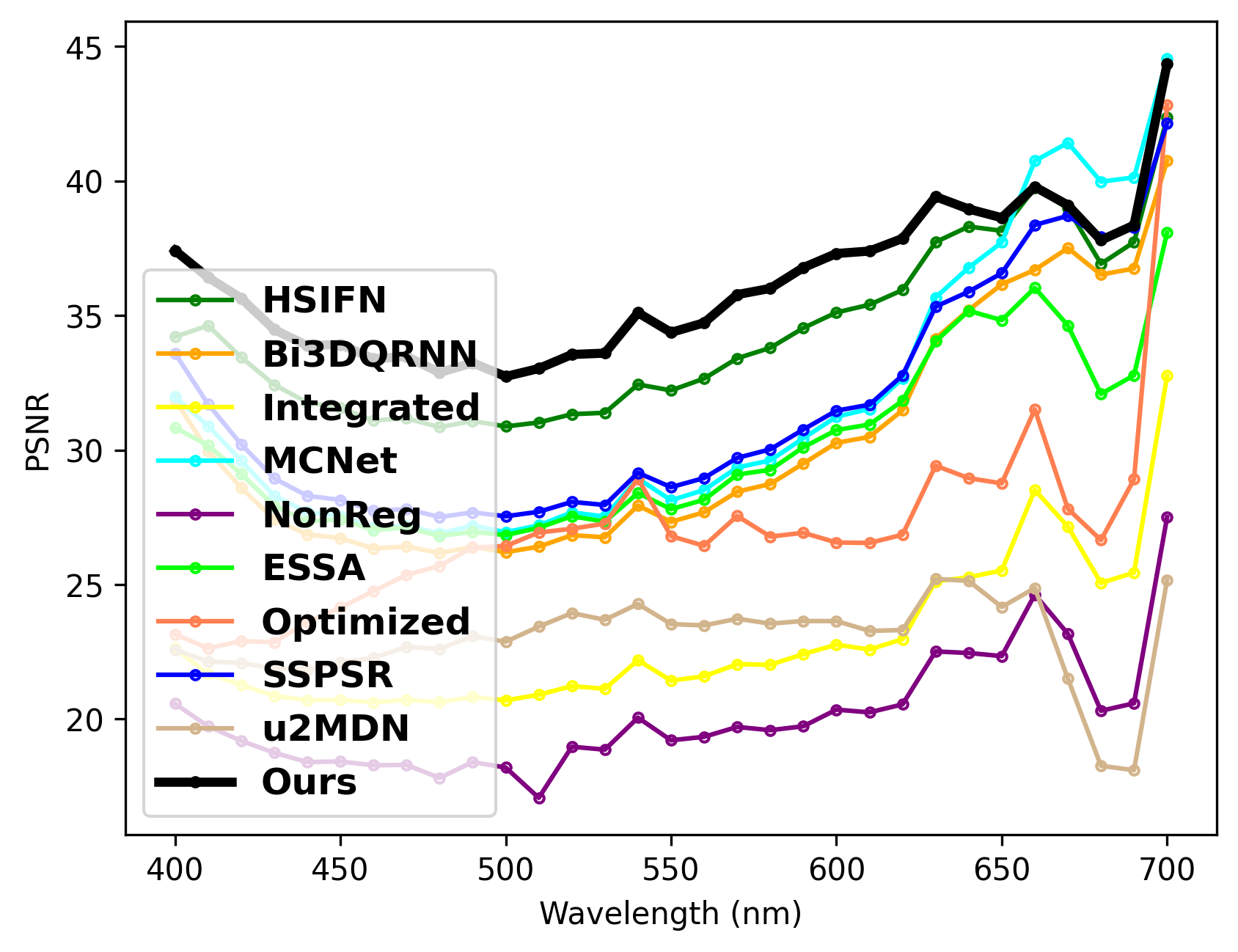}
          \caption{The curve of PSNR for each band.}
      \end{subfigure}
    \hfill
    \begin{subfigure}[h]{0.33\textwidth}
          \centering
          \small
          \includegraphics[width=1\linewidth]{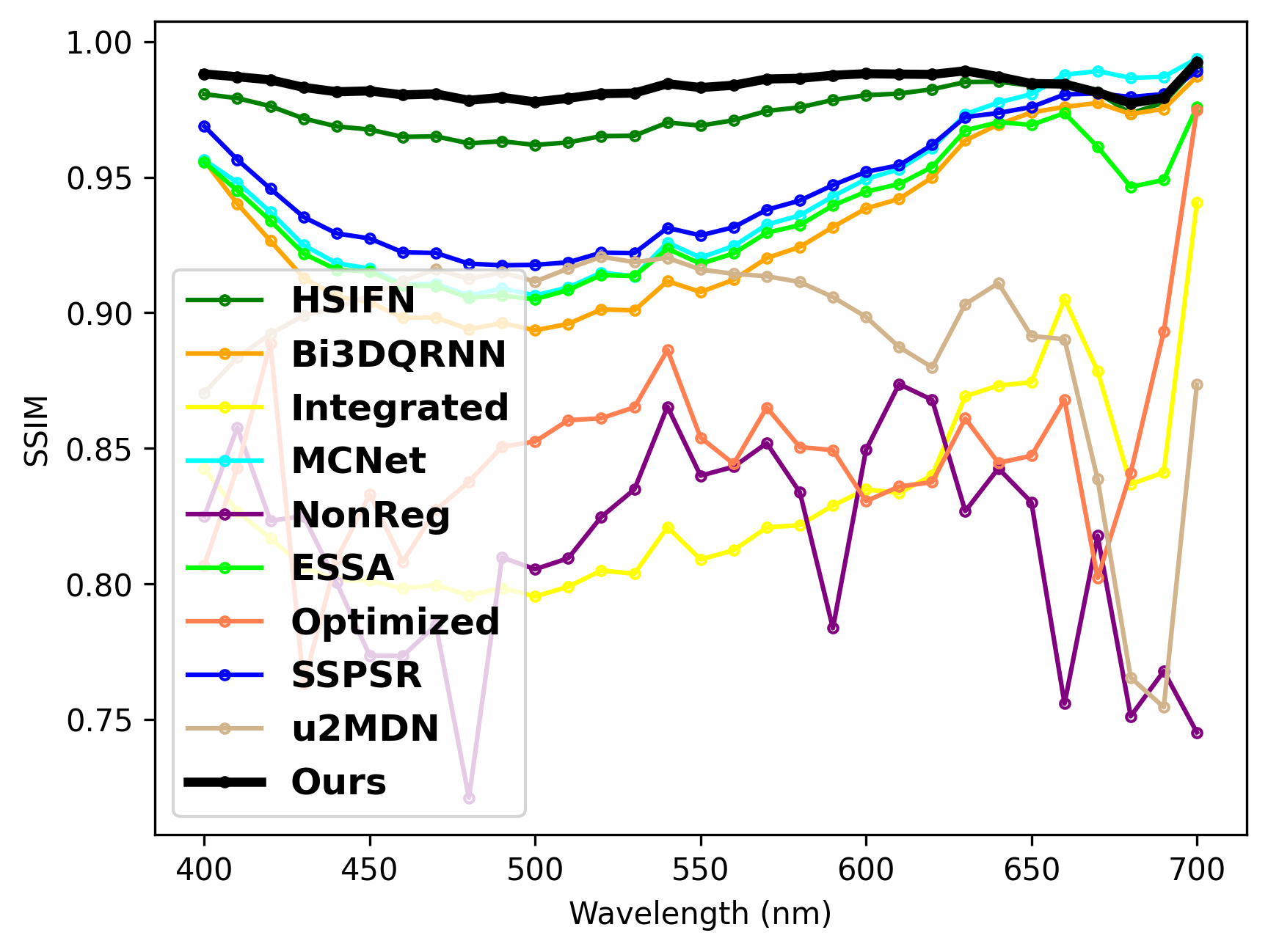}
          \caption{The curve of SSIM for each band.}
      \end{subfigure}
\end{center}
   \caption{Visualization of spectral fidelity and band-wise analysis. (a) The points of the spectral curve are in the center of the red box in Fig.~\ref{fig:RAFT-warp-warpNet}. It can be seen our result (Ours, black line) has the highest correlation coefficient with ground truth (red line). The PSNR curve and SSIM curve in (b) and (c) indicate that our method exhibits the highest performance in most spectral bands}
    \label{fig:Ablation spectral consistency}
\end{figure*}

\subsubsection{Effect of Attention Fusion Module}

We conduct further evaluations to investigate the effectiveness of the attention fusion (AF) module in Tables~\ref{tab:ablation-modules} and~\ref{tab:ablation-AF}. As shown in Table~\ref{tab:ablation-modules} last row, using the AF module gains 1.56 dB and 0.011 in PSNR and SSIM, which demonstrates the effectiveness of our proposed AF module. As shown in Table~\ref{tab:ablation-AF}, we provide a comparison with different components of AF. We first replace the attention-based architecture with UNet and maintain previous TSIA and FA modules. We then conduct two sets of comparison experiments, using only SSA (spectral-wise self-attention) to form the whole module and SSA+RCA (reference-guided cross-attention) to form the whole module respectively. When using the SSA module alone, we concatenate the features extracted from HSI and RGB to fusion for HR HSI.
It can be seen that using SSA gains 0.25 dB in PSNR and using combined  SSA and RCA gains 1.31 dB in PSNR over SSA alone, which indicates all these designs contribute to the performance improvement. 

\subsection{Discussion}

\noindent{\textbf{Model Size.}}
To further explore the effectiveness of our model, we compare it with different scale levels of the model in Table~\ref{tab:ablation-scales}. 
In our model, there are 3 levels of architecture, with feature channels increasing step by step. As shown in Fig.~\ref{fig:framework}, a new level is obtained after one downsampling in the encoder, and the combination of FA and AF yields a level. Thus, we obtain different scales of the network by reducing the above levels.
Our approach with the level of 3 can achieve PSNR, SSIM, and SAM of 37.37 dB, 0.974, and 0.042 at a scale factor of $\times8$ with a total number of 26.62 M parameters, which includes 5.26 M parameters for optical flow model RAFT. 
We also provide some suboptimal approaches with their parameters.
We can observe that our approach under level $L=2$ also outperforms HSIFN \citep{lai2024hyperspectral} with 21.01 M parameters and SSPSR \citep{jiang2020learning} with 20.82 M parameters, where our model has a smaller number of parameters, specifically 13.90 M.
We retain the single-level architecture of the network with a smaller number of parameters of 10.97 M, a computational cost of 165.6 G, and a single-image running time of 63.9 ms. 
It can be seen that our method surpasses all comparative methods, considering the number of parameters, FLOPs, and various performance metrics. These results validate the superior effectiveness of the proposed method.

\noindent{\textbf{Large Scale Factor.}}
In our experiments, we provide a large ratio of $\times16$ in addition to the common ratios of $\times4$ and $\times8$ in Section~\ref{sec: results} following HSIFN \citep{lai2024hyperspectral}. The maximum misalignment deviation in the dataset we use is below 30 pixels (\eg, Real dataset). For example, after downsampling for the original HSI the scale factor of $\times32$, the corresponding pixel deviation is less than 1 pixel. Therefore, due to the lack of spatial information in LR HSI itself and the small pixel deviation, it is difficult to align LR HSI and HR RGB explicitly.
As shown in Fig.~\ref{fig:large scale}, we sample the images using Gaussian blur and bicubic and set the scale factor to $\times4$, $\times8$, $\times16$, and $\times32$. We can observe that spatial details and relative positions can be distinguished at $\times4$ and $\times8$, but larger blurred textures are already present at $\times16$, and even the texture is no longer distinguishable at $\times32$.

\noindent{\textbf{Spectral Fidelity and Band-Wise Analysis.}}
To further analyze spectral fidelity and band-wise quality of different methods, we visualize the spectral curve of some selected pixels and the curves of PSNR and SSIM for each band. As shown in Fig.~\ref{fig:Ablation spectral consistency}, it can be seen the spectra recovered by our approach are more approximate to ground truth with a high correlation coefficient, which indicates that our method has higher spectral fidelity. In the curves of PSNR and SSIM for each band, we can observe that our method has suboptimal results in a few spectral bands, but overall we have the best performance.

\section{Conclusion}
\label{sec:conclusion}

In this paper, we propose an effective network for unaligned reference-based hyperspectral image super-resolution with spatial and spectral concordance. 
To address the alignment issue between LR HSI and HR RGB with resolution gap and distribution gap, we introduce a two-stage image alignment module with a synthetic generation pipeline for training. It relies on the fine-tuned optical flow model to generate a more accurate optical flow and warp model for texture detail refinement.
To enhance the interaction between alignment and fusion modules, we introduce an iterative deformable feature aggregation block in the feature aggregation module. It iteratively generates a learnable offset based on fusion results to achieve feature matching and texture aggregation. 
Furthermore, we introduce an attention fusion module based on spectral-wise self-attention and reference-guided cross-attention, which can explore the spectral correspondence and exhibit spectral concordance for the final HR HSI.

The experimental results on three HSI datasets, containing simulated/real natural datasets and remote-sensing datasets show that the proposed method outperforms many state-of-the-art methods under several comprehensive quantitative assessments. We have also conducted several ablation studies to analyze the advantages of our approach from different perspectives with quantitative metrics and visualization results.

We present a new effective paradigm for unaligned reference-based HSI super-resolution, but it still has some limitations. 
For example, in our data processing, we adopt the general degradation model that encompasses blurring and downsampling operations, but it might exhibit limited applicability when confronted with unique scenarios characterized by complex degradation patterns. 
In addition, the option to use more spatial texture information from multiple reference images could also be more beneficial for HSI super-resolution.

\noindent{\textbf{Data Availability.}} Data sharing does not apply to this article, as no datasets were generated or analyzed during the current study.

\noindent{\textbf{Acknowledgements.}} This work was supported by the National Natural Science Foundation of China (62331006, 6217 1038, 61931008, and 62088101).


%
%


\bibliographystyle{spbasic}
\bibliography{mybib}

\end{document}